\documentclass[journal]{IEEEtran}
\ifCLASSINFOpdf
\else
\fi
\usepackage[table,x11names]{xcolor}
\usepackage{soul}

\newcommand{\bigO}[1]{\mathcal{O}(#1)}

\newcommand{\mc}[1]{\mathcal{#1}}

\newcommand{\vts}[1]{\lvert #1 \rvert}
\newcommand{\Vts}[1]{\lVert #1 \rVert}
\newcommand{\bb}[1]{\mathbb{#1}}
\newcommand\inv[1]{#1\raisebox{1.05ex}{$\scriptscriptstyle-\!1$}}
\newcommand\Tstrut{\rule{0pt}{2.6ex}}         
\newcommand\Bstrut{\rule[-1.3ex]{0pt}{0pt}}   
\newcommand\Bstrutfrac{\rule[-0.7ex]{0pt}{0pt}}   
\newcommand\Tstrutfrac{\rule{0pt}{2.1ex}}         

\newcommand{\pc}{\zeta_c(t)}
\newcommand{\pd}{\zeta_d(t)}

\definecolor{sg}{HTML}{00ff7f}
\definecolor{lb}{HTML}{b0f5ef}
\definecolor{lg}{HTML}{9bfaa8}

\usepackage[most]{tcolorbox}
\newtcolorbox{highlighted}{colback=lg,coltext=black,breakable}
\makeatletter
\newcommand\footnoteref[1]{\protected@xdef\@thefnmark{\ref{#1}}\@footnotemark}
\makeatother

\newcommand{\shorteq}{%
  \settowidth{\@tempdima}{-}
  \resizebox{\@tempdima}{\height}{=}%
}

\usepackage{amssymb,fge}

\usepackage{amsmath, amssymb}
\usepackage{subcaption}
\usepackage[utf8]{inputenc}
\usepackage[T1]{fontenc}
\usepackage[english]{babel}
\usepackage[linesnumbered,ruled,vlined]{algorithm2e}
\usepackage[font=small,belowskip=-1.5pt]{caption} 
\usepackage{array}
\usepackage{graphicx}
\usepackage{amsfonts}
\usepackage{soul}
\usepackage{hhline}
\usepackage{multirow, makecell}
\usepackage{float}
\usepackage{booktabs}
\usepackage{anyfontsize}
\let\proof\relax

\usepackage{amsthm}
\usepackage{color}
\usepackage{transparent}
\usepackage{url}
\usepackage{footmisc}
\usepackage{setspace}
\usepackage[colorlinks=true, citecolor=magenta, breaklinks=true]{hyperref}
\usepackage{mathtools}
\usepackage{microtype}
\DeclarePairedDelimiter\abs{\lvert}{\rvert}

\DeclareMathOperator*{\argmax}{arg\,max}
\DeclareMathOperator*{\argmin}{arg\,min}
\theoremstyle{plain}

\newtheorem{definition}{Definition}[section]

\newtheorem{problem}{Problem}[section]
\DeclareMathOperator{\EX}{\mathbb{E}}

\begin{document}
%
\title{Using Graph-Theoretic Machine Learning to Predict Human Driver Behavior}
%
%
%

\author{Rohan~Chandra,
        Aniket~Bera,
        and~Dinesh~Manocha
\thanks{Manuscript received January 31, 2021; revised October 3, 2021. This work was supported in part by ARO Grants W911NF1910069 and W911NF1910315, Semiconductor Research Corporation (SRC), and Intel.}
\thanks{All authors are with the the Department
of Computer Science, University of Maryland, College Park,
MD, 20742, USA. e-mail: rchandr1@umd.edu.}
}

%
%

\markboth{}%
{Shell \MakeLowercase{\textit{et al.}}: Bare Demo of IEEEtran.cls for IEEE Journals}
%



\maketitle
\begin{abstract}%

Studies have shown that autonomous vehicles (AVs) behave conservatively in a traffic environment composed of human drivers and do not adapt to local conditions and socio-cultural norms. It is known that socially aware AVs can be designed if there exists a mechanism to understand the behaviors of human drivers. We present an approach that leverages machine learning to predict, the behaviors of human drivers. This is similar to how humans implicitly interpret the behaviors of drivers on the road, by only observing the trajectories of their vehicles. We use graph-theoretic tools to extract driver behavior features from the trajectories and machine learning to obtain a computational mapping between the extracted trajectory of a vehicle in traffic and the driver behaviors. Compared to prior approaches in this domain, we prove that our method is robust, general, and extendable to broad-ranging applications such as autonomous navigation. We evaluate our approach on real-world traffic datasets captured in the U.S., India, China, and Singapore, as well as in simulation.
\end{abstract}

\section{Introduction}
\label{sec: introduction}

Autonomous Vehicles (AVs) are an active area of research, successfully employing tools from machine learning~\cite{ml_av_review}, perception~\cite{perception_review}, planning and driver behavior modeling~\cite{schwarting2019social}. Recently, there have been multiple breakthroughs in  perception-based tasks in autonomous driving in areas that include object detection~\cite{guan2021m3detr}, tracking~\cite{chandra2019densepeds, chandra2019roadtrack}, trajectory prediction~\cite{chandra2019robusttp, chandra2019traphic, chandra2020forecasting}, and planning~\cite{chandra2021gameplan, bgap}. While these advances have been widely successful, current AVs still lack the ability to interact with multiple human drivers in dense traffic scenarios~\cite{chandra2021meteor} such as intersections and merging on highways.

As a precautionary measure, AVs are designed to behave conservatively in order to maximize safety~\cite{sa2}. A recent study~\cite{SHI2021100003} conducted real-world AV experiments and collected factors that may associate with how people's opinions change before and after experiencing a ride in an AV. However, conservative AV behavior is not always desirable or necessary, particularly given potential consequences like low efficiency and shortsighted behavior, which frequently frustrate other human drivers~\cite{dirtyTesla-gamma}. For example, in~\cite{dirtyTesla-gamma}, a Tesla driver is observed to be executing a lane-change maneuver. The Tesla AutoPilot slows down to wait for an excessively large gap in the target lane thereby blocking the traffic behind it in the current lane. This causes frustration and inefficiency among the blocked drivers. Furthermore, some studies~\cite{sa2} have pointed out that aggressive driving for AVs is even desirable in certain situations like reconnaissance, material transport, emergency handling, or efficiency-sensitive application. Sometimes it is even desirable simply based on human preference. 

There is prior work on predicting human driver behavior from trajectory data using machine learnings~\cite{rabinowitz2018machine, schwarting2019social, chandra2019graphrqi, cmetric, chandra2020stylepredict}. The two main approaches for this task include inverse reinforcement learning (IRL) and machine learning (regression and clustering techniques). Due to the data-driven nature of these methods, they incur two predominant limitations, which are inherent to most learning-based techniques in artificial intelligence research. First, these data-driven methods are constrained to a narrow range of traffic environments and fail to generalize to different environments. 
Furthermore, it has been shown both empirically and theoretically~\cite{nnrobust} that data-driven methods are not robust to fluctuations or noise in the sensor measurements (GPS, lidars, depth cameras etc.). These limitations prevent the current driver behavior prediction systems from being used in other AD tasks such as navigation.

Furthermore, in autonomous driving, it is important to handle the unpredictability and aggressive nature of human drivers during navigation. While there is considerable research on designing navigation algorithms~\cite{schwarting2018planning}, much of it assumes little to no interaction with human drivers. However, in real-life circumstances, drivers may act irrationally by moving in front of other vehicles, suddenly changing lanes, or aggressively overtaking. One such instance occurred in 2016 when an AV by Google collided with an oncoming bus during a lane change maneuver~\cite{davies2016google}. The AV assumed that the bus driver was going to yield; instead, the bus driver accelerated. Therefore, we need navigation methods that can account for different driver behaviors.



\textbf{Main Contributions:} In light of the limitations presented by data-driven methodologies, we present a fundamentally different approach to driver behavior prediction that can \textit{generalize} to widely varying traffic scenarios while also being \textit{robust} to realistic fluctuations in sensor noise. Our model not only alleviates the problems of prior approaches in order to predict human driver behavior, but also extends existing navigation research to behaviorally-guided navigation. Our main contributions include:

\begin{enumerate}
    \item A new approach to predict driver behavior from raw vehicle trajectories using graph-theoretic machine learning. In this approach, called StylePredict, we use the concept of vertex centrality functions~\cite{rodrigues2019network} and spectral analysis to measure the likelihood and intensity of driving styles such as overspeeding, overtaking, sudden lane-changes, etc. This process generates driver behavior features that can then be used for training machine learning algorithms.

    \item Extending current navigation research~\cite{schwarting2018planning} to \textit{behaviorally-guided} local navigation. This novel approach to navigation computes a local trajectory for the AV, taking into account the aggressiveness or conservativeness of human drivers. For example, the AV learns to slow down around aggressive human drivers while confidently overtaking conservative drivers. 

\end{enumerate}

StylePredict can be deployed in real-world traffic. We test extensively on real-world traffic datasets (Section~{\ref{subsec: datasets}}, Table~{\ref{tab: accuracy}}) collected in India, Singapore, U.S.A, and China. These datasets contain sensor noise (latency, precision, presence of outliers, etc.) identical to that expected in the real world. In Section~{\ref{subsec: noise_invariance}} and Table~{\ref{tab: sim_analysis}}, we demonstrate robustness of our method to these sensor issues. 
\section{Related Work}
\label{sec: related_work}

\subsection{Graph-based Machine Learning}
\label{subsec: graph_ml}

Graph-based machine learning is a sub-field in machine learning where the input data is organized as graphs. While the core learning algorithms themselves, including neural networks, LSTMs~\cite{lstm}, and convolutional neural networks, remain the same, they are now referred to as graph neural networks (GNNs)~\cite{gnn_review}, Graph-LSTMs~\cite{chandra2020forecasting}, and graph convolutional networks (GCNs)~\cite{gcn_review}, respectively. Graph-based machine learning algorithms have been widely used in trajectory prediction, computer vision and natural language processing~\cite{gnn_applications}. GNNs, Graph-LSTMs, and GCNs, however, are ``deep'' networks and require a huge amount of training data in order to produce meaningful results.

In this work, we instead use ``shallow'' graph-based machine learning, which includes learning algorithms based on logistic regression~\cite{park2015logistic}, multi-layer perceptrons, and support vector machines~\cite{svm}, which require fewer computational resources than deep learning-based methods.

\subsection{Data-Driven Methods for Driver Behavior Prediction}
\label{subsec: data_driven}

Data-driven methods broadly follow two approaches. In the first approach, various machine learning algorithms including clustering, regression, and classification are used to predict or classify the driver behavior as either aggressive or conservative. These methods have mostly been studied in traffic psychology and the social sciences~\cite{ernestref14,ernestref15,ernestref16}. So far, there has been relatively little work to improve the robustness and ability to generalize to different traffic scenarios, which require ideas from computer vision and robotics. In this work, we bridge the gap between robotics, computer vision, and the social sciences and develop an improved graph-theoretic machine learning model for human driver behavior prediction that alleviates the limitations of prior approaches.

The second approach uses trajectory data to learn reward functions for human behavior using inverse reinforcement learning (IRL)~\cite{rabinowitz2018machine, schwarting2019social, gt2}. IRL-based methods, however, have certain limitations. IRL requires large amounts of training data and the learned reward functions are unrealistically tailored towards scenarios only observed in the training data~\cite{rabinowitz2018machine, gt2}. For instance, the approach proposed in~\cite{rabinowitz2018machine} requires $32$ million data samples for optimum performance. Additionally, IRL-based methods are sensitive to noise in the trajectory data~\cite{schwarting2019social, gt2}. Consequently, current IRL-based methods are restricted to simple and sparse traffic conditions.

\subsection{Navigation Research in Autonomous Driving}
\label{subsec: Related_Work:Navigation_Algorithms}
Navigation in robotics is a well studied area of research. At a broad level, navigation methods can be categorized into approaches for vehicle control, motion planning, and end-to-end learning-based methods. Techniques for vehicular control methods assume apriori an accurate motion model of the vehicle. Such methods can be used for controlling vehicles at high speeds or during complex maneuvers. Motion planning methods can be further sub-divided into lattice-based~\cite{ferguson2008motion}, probabilistic search-based~\cite{lavalle2001randomized}, or use non-linear control optimization~\cite{liniger2015optimization} approaches. 

In addition to vehicular control and motion planning methods, many learning-based techniques are also used~\cite{jiapan2018crowdmove,jiapan2018fully,everett2018motion}. These methods are based on reinforcement learning where one finds an optimal policy that directly maps the sensor measurements to control commands such as velocity or acceleration and steering angle. Li et al.~\cite{li2019learning} formulate the navigation problem as one of action prediction using the proximity relationship between agents along with their visual features. 

However, the above methods do not consider the interaction among human drivers. Typically, in order to model dynamic obstacles, prior methods have either assumed a linear constant velocity model~\cite{everett2018motion}. 
Our behavior-based formulation can be integrated with these methods. We refer the reader to~\cite{schwarting2018planning} for a detailed review on recent planning and navigation methods.

\subsection{Interpretation of Driver Behavior in Social Science }
\label{subsec: driver_behavior}

Many studies have attempted to define driver behavior for traffic-agents. 
Sagberg et al.~\cite{sagberg2015review} extract and summarize the common elements from these definitions and propose a unified definition for driver behavior. We incorporate this definition in our driver behavior model.

\begin{definition}
\textbf{(Sagberg et al.~\cite{sagberg2015review}} Driver behavior refers to the high-level \textit{``global behavior''}, such as aggressive or conservative driving. Each global behavior can be expressed as a combination of one or more underlying \textit{``specific styles''}. For example, an aggressive driver (global behavior) may frequently overspeed or overtake (specific styles).
\label{def: behavior}
\end{definition}

\noindent The main benefit of Sagberg's definition is that it allows for a formal taxonomy for driver behavior classification. Specific indicators can be classified as either \textit{longitudinal} styles (along the axis of the road) or \textit{lateral} (perpendicular to the axis of the road). 
We can formally characterize driver behavior by mathematically modeling the underlying specific indicators.

\begin{problem}
In a traffic video with $N$ vehicles during any time-period $\Delta t$, given the trajectories of all vehicles, our objective is to \ul{mathematically model} the \textit{specific styles} for all drivers during $\Delta t$.
\label{problem: prob}
\end{problem}

\noindent In Section~\ref{sec: approach}, we elucidate on ``mathematically modeling'' a specific style. In Section~\ref{subsec: DGG}, we construct the ``traffic-graph'' data structure used by our approach. We introduce the ideas of vertex centrality in Section~\ref{sec: centrality} followed by a presentation of our main approach in Section~\ref{sec: approach}. We describe the experiments and results in Section~\ref{sec: experiments_and_results}. We conclude the paper in Section~\ref{sec: conclusion}.
\section{Representing Traffic Data Using Graphs}
\label{subsec: DGG}
The behavior of drivers depend on their interactions with nearby drivers. StylePredict models the relative interactions between drivers by representing traffic through weighted undirected graphs called ``traffic-graphs''. In this section, we describe the construction of these graph representations. If we assume that the trajectories of all the vehicles in the video are extracted using state-of-the-art localization methods~\cite{bresson2017simultaneous} and are provided to our algorithm as an input, then the traffic-graph, $\mc{G}_t$, at each time-step $t$ can be defined as follows,
\begin{figure*}
\centering
\includegraphics[width=\linewidth]{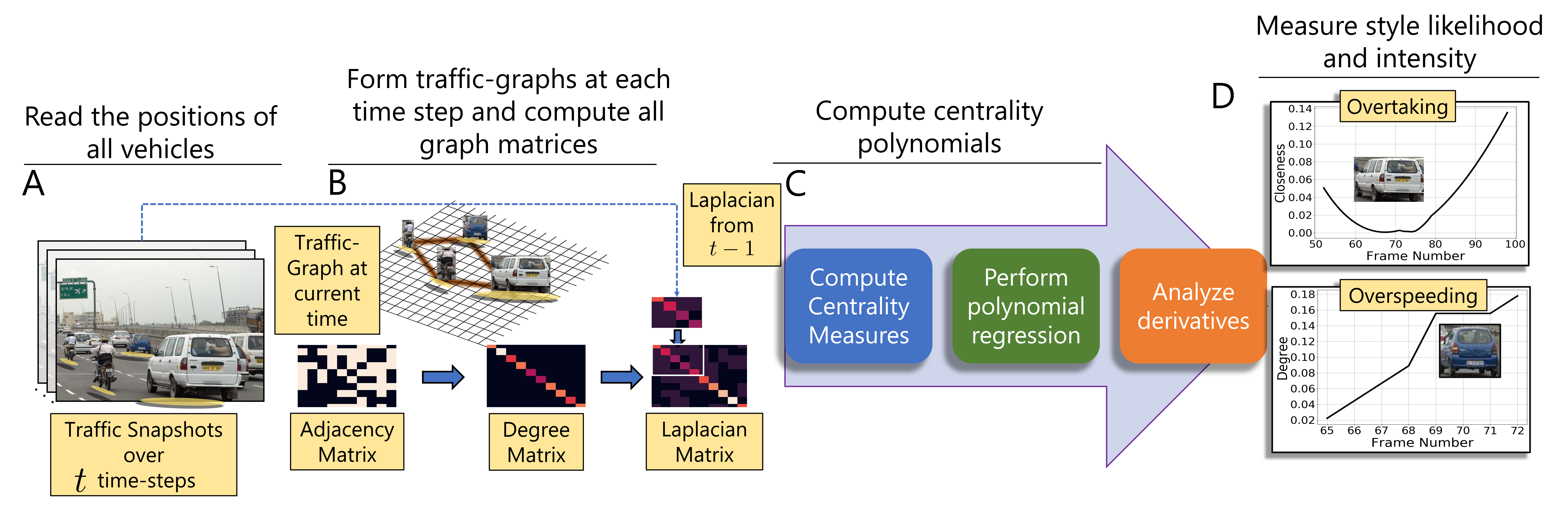}
\caption{\textbf{Overview:} The autonomous vehicle reads the positions of all vehicles in realtime. The positions and corresponding spatial distances between vehicles are represented through a traffic-graph $\mathcal{G}_t$ (Section~\ref{subsec: DGG}). We use the centrality functions defined in Section~\ref{sec: centrality} to model the specific driving style corresponding to the global behaviors as outlined in Table~\ref{tab: behaviors_centrality}.}
\label{fig: overview}
\end{figure*}
\begin{definition}
A ``traffic-graph'', $\mc{G}_t$, is a dynamic, undirected, and weighted graph with a set of vertices $\mathcal{V}(t)$ and a set of edges $\mc{E}(t) \subseteq \mc{V}(t) \times \mc{V}(t)$ as functions of time defined in the 2-D Euclidean metric space with metric function $f(x,y) = \Vts{x-y}^2$. Two vertices $v_i, v_j \in \mc{V}$ are connected if and only if $f(v_i,v_j) < \mu$, where $\mu$ is a distance threshold parameter.
\end{definition}

We use $N$ to represent the maximum number of vehicles tolerated by our system. $N$ is typically fixed as some large integer (e.g., $1000$) for each vehicle. Most real world commercial and academic systems use large high-performing computers to run computations involving large matrices~\cite{hpc}. Therefore, large values of $N$ do not impose a computational burden on our approach. Road-agents at each time instance $t$ in a traffic scenario can be represented using a traffic-graph $\mc{G}_t$. Each vertex in the graph $\mc{G}_t$ is represented by the vehicle position in the global coordinate frame, \textit{i.e.} $v_i \gets [x_i,y_i]^\top \in \mathbb{R}^2$. The spatial distance between two vehicles is assigned as the cost of the edge connecting the two vehicles.

In computational graph theory, every graph $\mc{G}$ can be equivalently represented by an adjacency matrix, $A$. For a particular traffic-graph $\mc{G}$, the adjacency matrix $A$ is given by $A(i,j)=(v_i,v_j)$ if $f(v_i,v_j) < \mu,i \neq j$ (otherwise $0$). 
Adjacency matrices allow linear vector operations to be performed on graph structures, which are useful for analyzing individual vertices. For example, each non-zero entry in the $j^\textrm{th}$ column corresponding to the $i^\textrm{th}$ row of the adjacency matrix stores the relative distance between the $i^\textrm{th}$ and $j^\textrm{th}$ vehicles. $A$ is initialized as an $N\times N$ identity matrix.



However, considering the traffic-graph and its corresponding adjacency matrix only at a current time-step $t$ is not useful in describing the behavior of a driver. The behavior of a driver also depends on their actions from previous time-steps. To accommodate this notion, at each time-step $t$, we populate $A$ with principle sub-matrices $A_t$ of size $t \times t$,

\[\underbrace{\begin{bmatrix}
   \overbrace{\begin{bmatrix}
              \overbrace{\begin{bmatrix}
             \mathbf{A_1} & a_{12} \\
             a_{21} & a_{22} \\
    \end{bmatrix}}^{\mathbf{A_2}} &  \begin{matrix}
     \mathbf{a_{13}} \\ \mathbf{a_{23}} \\
    \end{matrix}\\
             \begin{matrix}
     \mathbf{a_{31}} &\mathbf{a_{32}} \\
    \end{matrix}  & \mathbf{1}\\
    \end{bmatrix}}^{\mathbf{A_3}}& \begin{matrix}
    \dots  & 0 \\
    \end{matrix} \\
    \begin{matrix}
     \vdots  \\ 0 \\
    \end{matrix} & \begin{matrix}
              \ddots & \vdots\\
              \dots & 1 \\
    \end{matrix} \\
\end{bmatrix}}_{\mathbf{A_{N \times N}}}.\]

\noindent The sub-matrix for the next time-step, $A_{t+1}$, is obtained by the following update,

\begin{equation}
A_{t+1} =
\overbrace{\left[
\begin{array}{c|c}
    [A_{t}]_{\mathbf{t \times t}} & 0 \\
\hline
0  & 1
\end{array}
\right]}^{\mathbf{(t+1) \times (t+1)}} + \ \delta  \sigma (\delta)^\top,
\label{eq: A_update}
\end{equation}
\noindent where $\delta  \sigma (\delta)^\top \in \bb{R}^{(t+1) \times (t+1)} $ is a sparse update matrix and $\delta, \sigma(\delta)$ are update vectors defined as follows,

\[\delta = \begin{bmatrix}
\overbrace{\delta_{11}}^{\mathbf{\delta_{11}\neq 0 }} & 0 \\
\delta_{21} & 0 \\
\vdots & \vdots\\
\delta_{t1} & 0 \\
0 & 1 \\
\end{bmatrix}_{\mathbf{(t+1) \times 2}}
\sigma(\delta) = \begin{bmatrix}
0 & \overbrace{\delta_{11}}^{\mathbf{\delta_{11}\neq 0 }} \\
0 & \delta_{21} \\
\vdots & \vdots\\
0 & \delta_{t1} \\
1 & 0 \\
\end{bmatrix}_{\mathbf{(t+1) \times 2}}.\]

\noindent Here, $\sigma$ is a permutation that swaps the two columns in $\delta$. If the $j^\textrm{th}$ row of $\delta$ is non-zero, then that implies that the $j^\textrm{th}$ road-agent has formed a new edge with a new vehicle that came into its proximity; this new vehicle will be added to the current traffic-graph. This new vehicle is identified by a unique ID number provided by the localization sensor (GPS or lidar). For example, if a vehicle with ID $1 (j=1)$ has formed a new edge connection with another vehicle. This corresponds to $\delta_{11} \neq 0$. The update rule in Equation~{\ref{eq: A_update}} ensures that a vehicle adds edge connections to new vehicles while retaining edge connections with previously seen vehicles.


\begin{figure*}[t]
\centering
  \begin{subfigure}[h]{\textwidth}
    \includegraphics[width=\textwidth]{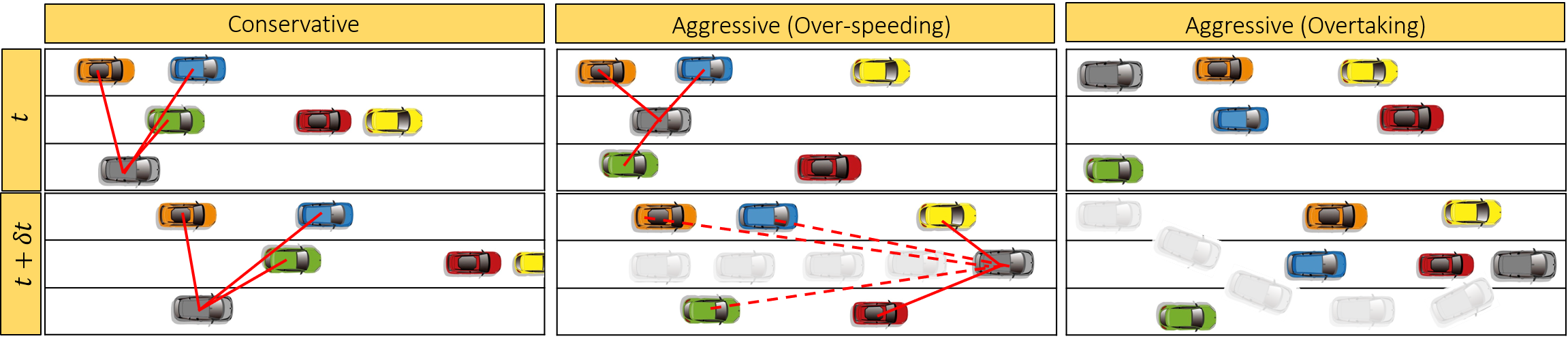}
    \caption{In all three scenarios, the ego-vehicle is a gray vehicle marked with a blue outline. (\textit{left}) A conservative vehicle, \textit{(middle)} an overspeeding vehicle in the same lane, and \textit{(right)} a weaving and overtaking vehicle. }
    \label{fig: part_a}
  \end{subfigure}
  \begin{subfigure}[h]{0.3\textwidth}
    \includegraphics[width=\textwidth]{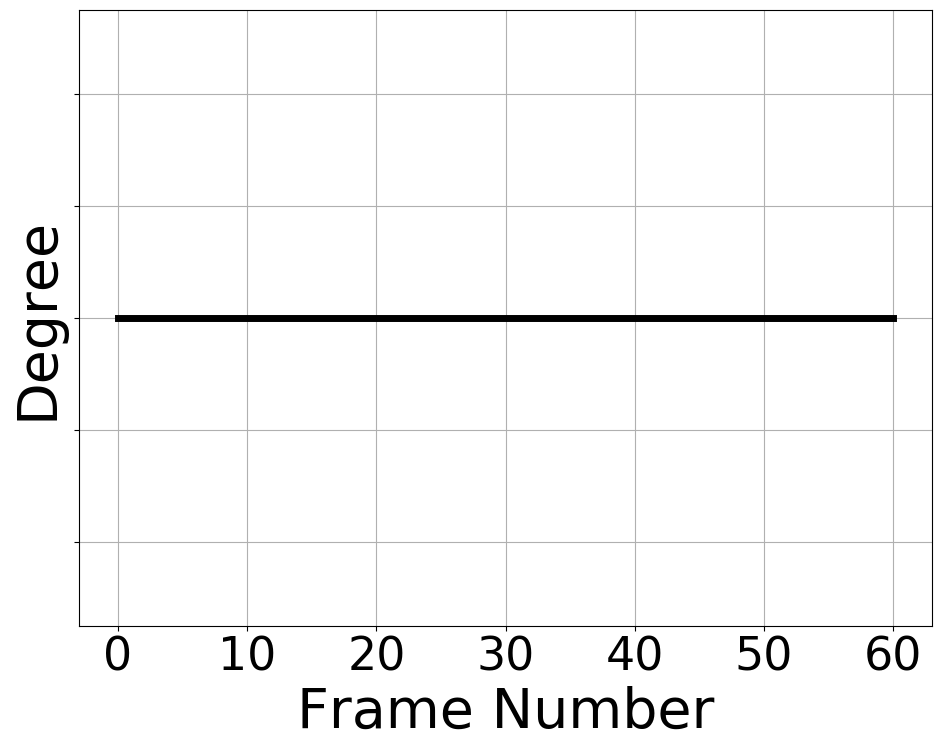}
    \caption{Constant degree centrality function for conservative vehicle.}
    \label{fig: part_b}
  \end{subfigure}
    \begin{subfigure}[h]{0.3\textwidth}
    \includegraphics[width=\textwidth]{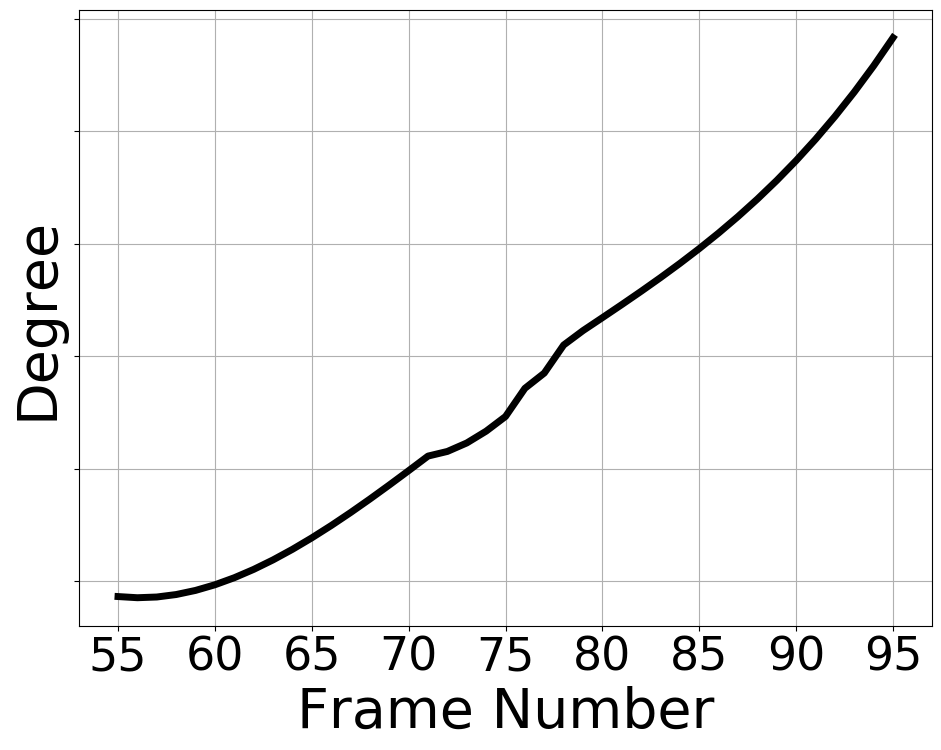}
    \caption{Monotonically increasing centrality function for overspeeding vehicle.}
    \label{fig: part_c}
  \end{subfigure}
  \begin{subfigure}[h]{0.3\textwidth}
    \includegraphics[width=\textwidth]{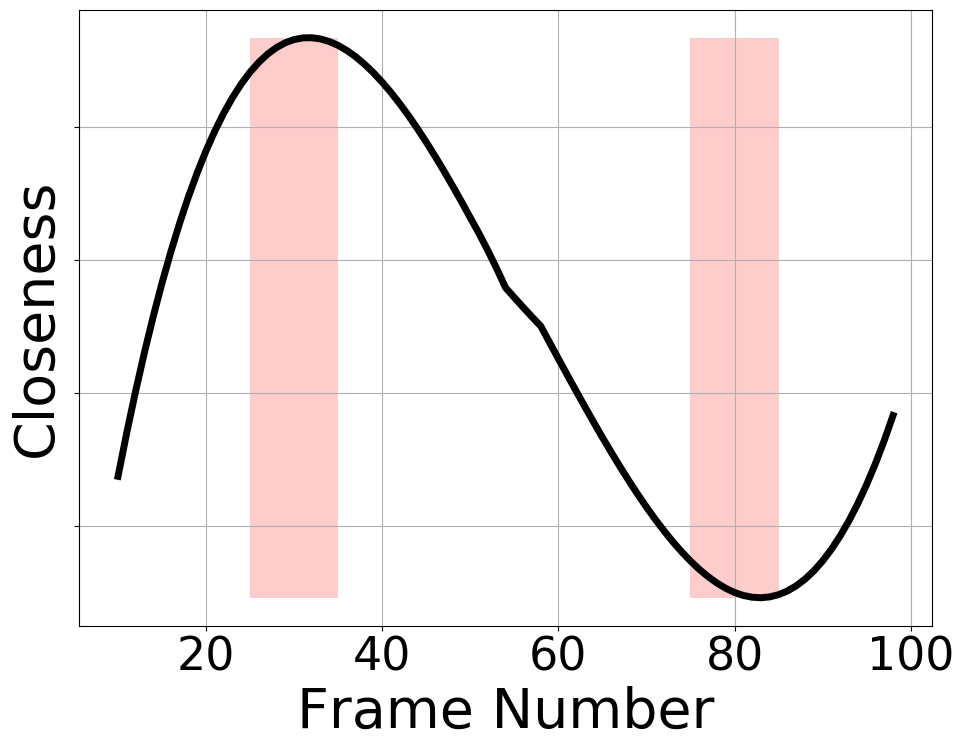}
    \caption{Extreme points for closeness centrality function for weaving vehicle.}
    \label{fig: part_d}
  \end{subfigure}
  \caption{\textbf{Measuring the Likelihood of Specific Styles:} We measure (degree and closeness centrality) the likelihood that an ego-vehicle (grey with a blue outline) has a specific driving style by computing the magnitude of the derivative of the centrality functions as well as the functions' extreme points. In Figure~\ref{fig: part_b}, the derivative of the degree centrality function is $0$ because the ego-vehicle does not observe any additional new neighbors (See Section~\ref{subsec: SLE_SIE}), so the degree centrality is a constant function; therefore, the vehicle is conservative. In Figure~\ref{fig: part_c}, the vehicle overspeeds and, consequently, the rate of observing new neighbors is high, which is reflected in the magnitude of the derivative of the degree centrality being positive. Finally, in Figure~\ref{fig: part_d}, the ego-vehicle demonstrates overtaking/sudden lane-changes and weaves through traffic. These behaviors are reflected in the magnitude of the slope and the location of extreme points, respectively, of the closeness centrality function.}
  \label{fig: maneuver_behaviors}
\end{figure*}

A candidate vehicle is categorized as ``new'' with respect to a vehicle if there does not exist any prior edge connection between the vehicles \textit{and} the speed of the old vehicle is greater than the candidate vehicle. If an edge connection already exists between the vehicle, then the candidate vehicle is said to have been ``observed'' or ``seen''. The dimension of $A$ is constant ($N \times N$). Once the upper limit $N$ has been achieved ($N$ different vehicles have been observed), then $A$ is re-initialized as an $N \times N$ identity matrix. As the behaviors of each vehicle are determined in an online manner, ``erasing'' the old vehicles from the matrix to make way for new vehicles does not affect their behavior computation; their behaviors have already been computed and stored. If the number of vehicles is less than $N$, then the ``unused'' entries in $A$ are simply left as $0$.  Finally, vehicles appearing and disappearing from the field of view of the ego-vehicle does not impact the size of $A$. If a vehicle does not remain in the field of view of the ego-vehicle for a sufficient amount of time, then our algorithm does not consider that vehicle in the adjacency matrix $A$.

\section{StylePredict: Mapping Trajectories to Behavior}
\label{sec: stylepredict}

\subsection{Centrality Measures}
\label{sec: centrality}
\begin{table}[t]
\centering
\caption{Definition and categorization of driving behaviors~\cite{sagberg2015review}. We measure the likelihood and intensity of specific styles by analyzing the first-and second-order derivatives of the centrality polynomials.}
\centering
\resizebox{\columnwidth}{!}{
\begin{tabular}{lcccc}

\toprule

Global & Specific & Centrality & SLE & SIE\\
\midrule
 \multirow{3}{*}{Aggressive}& Overspeeding & Degree ($\zeta_d$) &   |$1^\textrm{st}$ Derivative| & |$2^\textrm{nd}$ Derivative|\\
 & Overtaking / SLC & Closeness ($\zeta_c$) &  |$1^\textrm{st}$ Derivative|  &   |$2^\textrm{nd}$ Derivative|\\
  & Weaving  & Closeness ($\zeta_c$) &  Extreme Points & $\varepsilon$-sharpness \\
 \midrule

\multirow{2}{*}{Conservative}& Driving Slowly & Degree ($\zeta_d$)& |$1^\textrm{st}$ Derivative|& |$2^\textrm{nd}$ Derivative|\\
 & No Lane-change  & Closeness ($\zeta_c$)&|$1^\textrm{st}$ Derivative|& |$2^\textrm{nd}$ Derivative|\\

\bottomrule
\end{tabular}
}
\label{tab: behaviors_centrality}
\end{table}

In graph theory and network analysis, centrality measures~\cite{rodrigues2019network} are real-valued functions $\zeta:\mathcal{V}\longrightarrow \mathbb{R}$, where $\mathcal{V}$ denotes the set of vertices and $\mathbb{R}$ denotes a scalar real number that identifies key vertices within a graph network. So far, centrality functions have been restricted to identifying influential personalities in online social media networks~\cite{cen-socialmedia} and key infrastructure nodes in the Internet~\cite{cen-internet}, to rank web-pages in search engines~\cite{cen-pagerank}, and to discover the origin of epidemics~\cite{cen-epidemic}. There are several types of centrality functions. The ones that are of particular importance to us are the degree centrality and the closeness centrality denoted as $\pd$ and $\pc$, respectively. These centrality measures are defined in~\cite{cmetric} (See section III-C). Each function measures a different property of a vertex. Typically, the choice of selecting a centrality function depends on the current application at hand. In this work, the closeness centrality and the degree centrality functions measure the likelihood and intensity of specific driving styles such as overspeeding, overtaking, sudden lane-changes, and weaving~\cite{cmetric}.

 

\subsection{Algorithm}
\label{sec: approach}

Here, we present the main algorithm, called \textit{StylePredict}, for solving Problem~\ref{problem: prob}. StylePredict maps vehicle trajectories to specific styles by computing the likelihood and intensity of the latter using the definitions of the centrality functions. The specific styles are then used to assign global behaviors~\cite{sagberg2015review} according to Table~\ref{tab: behaviors_centrality}. We summarize the StylePredict algorithm as follows:


\begin{enumerate}
    \item Obtain the positions of all vehicles using localization sensors deployed on the autonomous vehicle and form traffic-graphs at each time-step (Section~\ref{subsec: DGG}).
    
    
    \item Compute the closeness and degree centrality function values for each vehicle at every time-step.
    
    \item Perform polynomial regression to generate uni-variate polynomials of the centralities as a function of time. 
    
    \item Measure likelihood and intensity of a specific style for each vehicle by analyzing the first- and second-order derivatives of their centrality polynomials.
    
    \item Classify the centrality polynomials, obtained from step $3$, as either aggressive or conservative using machine learning algorithms such as Multi-Layer Perceptrons (MLPs).
    
\end{enumerate}

\noindent We depict the overall approach in Figure~\ref{fig: overview}. We begin by using the construction described in Section~\ref{subsec: DGG} to form the traffic-graphs for each frame and use the definitions in~\cite{cmetric} to compute the discrete-valued centrality measures. Since centrality measures are discrete functions, we perform polynomial regression using regularized Ordinary Least Squares (OLS) solvers to transform the two centrality functions into continuous polynomials, $\pc$ and $\pd$, as a function of time. We describe polynomial regression in detail in the following subsections. We compute the likelihood and intensity of specific styles by analyzing the first- and second-order derivatives of $\pc$ and $\pd$ (this step is discussed in further detail in Section~\ref{subsec: SLE_SIE}).



\subsection{Polynomial Regression}
\label{subsec: polynomial_regression}
 
In order to study the behavior of the centrality functions with respect to how they change with time, we convert the discrete-valued $\zeta[t]$ into continuous-valued polynomials $\zeta(t)$, using which we calculate the first- and second-order derivatives of the centrality functions as explained in Section~\ref{subsec: SLE_SIE}.

In this work, we choose a quadratic\footnote{A polynomial with degree $2$.} centrality polynomial can be expressed as $\zeta(t) = \beta_0 + \beta_1 t + \beta_2t^2$, as a function of time. Here, $\beta = [\beta_0 \ \beta_1 \ \beta_2]^\top$ are the polynomial coefficients. These coefficients can be computed using ordinary least squares (OLS) equation as follows,

\begin{equation}
        \beta = \inv{(M^\top M)}M^\top \zeta^i
    \label{eq: noiseless_OLS}
\end{equation}

\noindent Here, $M \in \mathbb{R}^{T\times(d+1)}$ is the Vandermonde matrix~\cite{manocha1992algebraic}.
and is given by, 

\[
M = \begin{bmatrix}
    1 & t_1 & t^2_1 & \dots  & t^d_1 \\
    1 & t_2 & t^2_2 & \dots  & t^d_2 \\
    \vdots & \vdots & \vdots & \ddots & \vdots \\
    1 & t_T & t^2_T & \dots  & t^d_T
\end{bmatrix}
\]



\subsection{Style Likelihood and Intensity Estimates}
\label{subsec: SLE_SIE}
In the previous sections, we used polynomial regression on the centrality functions to compute centrality polynomials. In this section, we analyze and discuss the first and second derivatives of the degree centrality, $\pd$, and closeness centrality, $\pc$, polynomials. Based on this analysis, which may vary for each specific style, we compute the Style Likelihood Estimate (SLE) and Style Intensity Estimate (SIE)~\cite{cmetric}, which are used to measure the probability and the intensity of a specific style.

\paragraph{Overtaking/Sudden Lane-Changes}
Overtaking is when one vehicle drives past another vehicle in the same or an adjacent lane, but in the same direction. The closeness centrality increases as the vehicle navigates towards the center and vice-versa. The SLE of overtaking can be computed by measuring the first derivative of the closeness centrality polynomial using $\textrm{SLE}(t) = \abs*{\frac{\partial \pc}{\partial t}}$. The maximum likelihood $\textrm{SLE}_\textrm{max}$ can be computed as $\textrm{SLE}_{\textrm{max}} = \max_{t \in \Delta t}{\textrm{SLE}}(t)$. The SIE of overtaking is computed by simply measuring the second derivative of the closeness centrality using~$ \textrm{SIE}(t) = \abs*{\frac{\partial^2 \pc}{\partial t^2}}$. Sudden lane-changes follow a similar maneuver to overtaking and therefore can be modeled using the same equations used to model overtaking. 

\paragraph{Overspeeding}

The degree centrality can be used to model overspeeding. As $A_t$ is formed by adding rows and columns to $A_{t-1}$ (See Equation~\ref{eq: A_update}), the degree of the $i^\textrm{th}$ vehicle (denoted as $\theta_i$) is calculated by simply counting the number of non-zero entries in the $i^\textrm{th}$ row of $A_t$. Intuitively, a drivers that are overspeeding will observe new neighbors along the way (increasing degree) at a higher rate than conservative, or even neutral, drivers. Let the rate of increase of $\theta_i$ be denoted as $\theta_i^{'}$. By definition of the degree centrality and construction of $A_t$, the degree centrality for an aggressively overspeeding vehicle will monotonically increase. Conversely, the degree centrality for a conservative vehicle driving at a uniform speed or braking often at unconventional spots such as green light intersections will be relatively flat. 
Therefore, the likelihood of overspeeding can be measured by computing,
\[\textrm{SLE}(t) = \abs*{\frac{\partial \pd}{\partial t}}\]

\noindent Similar to overtaking, the maximum likelihood estimate is given by $\textrm{SLE}_{\textrm{max}} = \max_{t  \in \Delta t}{\textrm{SLE}}(t)$. Figures~\ref{fig: part_b} and~\ref{fig: part_c} visualize how the degree centrality can distinguish between an overspeeding vehicle and a vehicle driving at a uniform speed. 

\paragraph{Weaving} A vehicle is said to be weaving when it ``zig-zags'' through traffic. Weaving is characterized by oscillation in the closeness centrality values between low values towards the sides of the road and high values in the center. Mathematically, weaving is more likely to occur near the critical points (points at which the function has a local minimum or maximum) of the closeness centrality polynomial. The critical points $t_c$ belong to the set $\mathcal{T} =\big\{ t_c \big | \frac{\partial \zeta_c(t_c)}{\partial t} = 0 \big\}$. Note that $\mathcal{T}$ also includes time-instances corresponding to the domain of constant functions that characterize conservative behavior. We disregard these points by restricting the set membership of $\mathcal{T}$ to only include those points $t_c$ whose $\varepsilon-$sharpness~\cite{dinh2017sharp} of the closeness centrality is non-zero. The set $\mathcal{T}$ is reformulated as follows,
\begin{equation}
    \begin{aligned}
    \mathcal{T} = \Bigg \{ t_c \bigg | \frac{\partial \zeta_c(t_c)}{\partial t} = 0 \Bigg \} \\
    \textrm{s.t.} \max_{t \in \mathcal{B}_\varepsilon(t_c)} \frac{\partial \pc}{\partial t}  \neq \frac{\partial \zeta_c(t_c)}{\partial t}\\
    \end{aligned}
    \label{eq: weaving}
\end{equation}
\noindent where $\mathcal{B}_\varepsilon (y) \in \mathbb{R}^d$ is the unit ball centered around a point $y$ with radius $\varepsilon$. The SLE of a weaving vehicle is represented by $\vts{\mathcal{T}}$, which represents the number of elements in $\mathcal{T}$. The $\textrm{SIE}(t)$ is computed by measuring the $\varepsilon-$sharpness value of each $t_c \in \mathcal{T}$. Figure~\ref{fig: part_d} visualizes how the degree centrality can distinguish between an overspeeding vehicle and a vehicle driving at a uniform speed. 
\paragraph{Conservative Vehicles}

Conservative vehicles, on the other hand, are not inclined towards aggressive maneuvers such as sudden lane-changes, overspeeding, or weaving. Rather, they tend to stick to a single lane~\cite{ahmed1999modeling} as much as possible, and drive at a uniform speed~\cite{sagberg2015review} below the speed limit. Correspondingly, the values of the closeness and degree centrality functions in the case of conservative vehicles remain constant. Mathematically, the first derivative of constant polynomials is $0$. The SLE of conservative behavior is therefore observed to be approximately equal to $0$. Additionally, the likelihood that a vehicle drives uniformly in a single lane during time-period $\Delta t$ is higher when,

\[\abs*{\frac{\partial \pc}{\partial t}} \approx 0 \  \textrm{and} \ \max_{t \in \mathcal{B}_\varepsilon(t^{*})} \textrm{SLE}(t) \approx \textrm{SLE}(t_c) .\]

\noindent The intensity of such maneuvers will be low and is reflected in the lower values for the SIE. 

\subsection{Robustness to Noise}
\label{subsec: noise_invariance}

In the formulation above, our algorithm assumes perfect sensor measurements of the global coordinates of all vehicles. However, in real-world systems, even state-of-the-art methods for vehicle localization incur some measurement errors. We consider the case in which the raw sensor data is corrupted by some noise $\epsilon$. Without loss of generality, we prove robustness to noise for the degree centrality. Further, the analysis can be extended to other centrality functions. The discrete-valued centrality vector for the $i^\textrm{th}$ agent is given by $\zeta^i \in \mathbb{R}^{T\times 1}$. Therefore, $\zeta^1[2]$ corresponds to the degree centrality value of the $1^\textrm{st}$ agent at $t=2$. 

In the previous section, we showed that a noiseless estimator may be obtained by solving an ordinary least squares (OLS) system given by Equation~\ref{eq: noiseless_OLS}. However, in the presence of noise $\epsilon$, the OLS system described in Equation~\ref{eq: noiseless_OLS} is modified as,
\begin{equation}
        \tilde\beta = \inv{(M^\top M)}M^\top \tilde\zeta^i
    \label{eq: noisy_OLS}
\end{equation}
\noindent where $\tilde \zeta^i = \zeta^i + \epsilon$. Then we can prove that $\Vts{\tilde \beta - \beta} =  \bigO{\epsilon}$. We defer the proof to the supplementary material.

\subsection{Behavior Classification Using Machine Learning}
\label{subsec: behavior_classification}

We treat the centrality polynomials computed in Section~\ref{subsec: polynomial_regression} as features in a supervised learning paradigm. While our formulation is such that any classification algorithm can be used, we select Multi-Layer Perceptron (MLP) as the classification model due to its superior performance. We defer a comparison between different ML algorithms to Section~\ref{sec: experiments_and_results}.

Formally, let $\Phi$ denote the MLP model that takes in a centrality feature vector, $\zeta(t)$, as input and produces a $1$-hot vector encoding, $\hat y = \Phi(\zeta(t))$, of the behavior prediction as output. Let $y$ denote the corresponding ground-truth label for that agent. Then, for $N$ agents, a loss function can be framed as follows,

\begin{equation}
    \mathcal{L} (\theta) = \sum_{i=1}^N \Vts{y_i - \Phi(\zeta^i(t))}^2
    \label{eq: cost_function}
\end{equation}

\noindent where $\theta$ denote the MLP model parameters. The goal of the classification problem is to find the optimum values of $\theta$, say $\theta^{*}$, that minimizes Equation~\ref{eq: cost_function}. More simply,

\[\theta^{*} = \argmin_\theta \mathcal{L}(\theta)\]

\section{Experiments and Results}
\label{sec: experiments_and_results}
We begin with a discussion of the evaluation metrics, the Time Deviation Error (TDE) and the weighted classification accuracy, for validating and measuring the accuracy of behavior prediction methods in Section~\ref{subsec: TDE_metric}. Then, we describe the real-world traffic datasets and simulation environment used for testing our approach and outline the annotation algorithm used to generate ground-truth labels for aggressive and conservative vehicles in Section~\ref{subsec: datasets}. We use the TDE to validate our approach and analyze the results in real-world traffic datasets in Section~\ref{subsec: real_world_traffic_analysis}. Finally, we analyze the weighted accuracy of StylePredict and compare with state-of-the-art graph classification and behavior prediction methods in Section~\ref{subsec: behaviore_prediction_eval}.

\subsection{Evaluation Metrics}
\label{subsec: TDE_metric}
\begin{enumerate}
    \item Time Deviation Error (TDE)~\cite{cmetric}: We use the TDE to validate our approach to modeling driver behavior using StylePredict. The TDE measures the temporal difference between the moments when a human identifies a behavior and when that same behavior is modeled using StylePredict. 
    The TDE is given by the following equation,

\begin{equation}
    \textrm{TDE}_\textrm{style} =   \abs*{\frac{t_\textrm{SLE} - \EX[T]}{f}} 
    \label{eq: TDE}
\end{equation}

\noindent where $\EX$ denotes the expected time-stamp of an exhibited behavior in the ground-truth annotated by a human and $f$ is the frame rate of the video. $t_\textrm{SLE}$ is obtained using $\argmax_{t \in \Delta t}{\textrm{SLE}(t)}$ as explained in Section~\ref{sec: approach}, $\EX[T]$ is computed using Algorithm~\ref{alg: ex}, described in the following section.

\item Weighted Classification Accuracy: To measure the accuracy of StylePredict in predicting future behaviors, we report a weighted classification accuracy, which is defined as the fraction of correctly predicted behaviors, weighted by class frequencies
.

\end{enumerate}

\subsection{Datasets and Simulation Environment}
\label{subsec: datasets}

 Our testing environments consist of both simulation and real-world trajectory data. The simulation includes a top-view of the traffic while the trajectory data has been captured from front-view vehicle-based sensors. Both settings are the same in the sense that they provide the same type of information --  the coordinates of each vehicle with respect to a fixed frame of reference (camera center in the case of top-view or the ego-vehicle in the case of front view).

\paragraph{Simulation Environment} 

We use the Highway-Env simulator~\cite{leurent2019social} developed using PyGame. The simulator consists of a 2D environment where vehicles are made to drive along a multi-lane highway using the Bicycle Kinematic Model~\cite{polack2017kinematic} as the underlying motion model where the linear acceleration model is based on the Intelligent Driver Model (IDM)~\cite{treiber2000congested} and the lane changing behavior is based on the MOBIL~\cite{kesting2007general} model. We note here that more sophisticated car models such as the Ackermann steering model may be used. While many popular vehicle simulators~\mbox{\cite{carla, sumo}} do provide the option of Ackermann modeling, these simulators do not provide the behavior-rich environment needed for testing our algorithm. Therefore, we restrict ourselves to~\mbox{\cite{leurent2019social}} that can generate aggressive and conservative driver behaviors. Furthermore, most simulators that use Ackermann steering do so for modeling safety by preventing slipping of the tires during tight turns such as U-turns or intersection turns. Since our environment consists of a straight road with no turns, the Ackermann model holds little advantage over the Bicycle kinematic model in our case.



\begin{algorithm}[t]
    \SetKwInOut{Input}{Input}
    \SetKwInOut{Output}{Output}
\SetKwComment{Comment}{$\triangleright$\ }{}
\SetAlgoLined
\Input{$M$ participants, set of starting frames $S = \{s_1, s_2, \ldots, s_M\}$, set of ending frames $E = \{e_1, e_2, \ldots, e_M\}$ }
\Output{$\EX[T]$ for a video}
$s^* = \min S$\\
$e^* = \max E$\\
Initialize a counter $c_t = 0$ for each frame $t \in [s^*, e^*]$\\
\For {$t \in [s^*, e^*]$}{
\If{$t \in [s_m, e_m]$}{
$c_t \gets c_t + 1$
}
$\mathcal{P}(T=t) = c_t $
}
$\EX[T] = \sum_t tc_t$, $t = s^*, s^*+1, \ldots, e^*$
\caption{Computing $\EX[T]$ for each video in a dataset.}
\label{alg: ex}
\end{algorithm}
\paragraph{Real-World Datasets}

We have evaluated StylePredict on traffic data collected from geographically diverse regions of the world. In particular, we use data collected in Pittsburgh (U.S.A)~\cite{Argoverse}, New Delhi (India)~\cite{chandra2019traphic}, Beijing (China)~\cite{wang2019apolloscape}, and Singapore (private dataset).
The format of the data includes the timestamp, road-agent I.D., road-agent type, and spatial coordinates obtained via GPS or lidars. We understand that the characteristics of drivers in a particular city may not mirror those in other cities of the same country. Therefore, all results presented in this work correspond to the traffic in the specific city where the dataset is recorded.

\begin{table}[t]
\caption{\textbf{Analysing Simulation Results using TDE:} We analyze StylePredict by varying the traffic density, number of lanes, and the noise parameter $\epsilon$ (Equation~{\ref{eq: noisy_OLS}}). We observe that TDE increases as these parameters increase in value. We discuss these results in detail in Section~{\ref{subsec: real_world_traffic_analysis}}.}
\centering
\resizebox{\columnwidth}{!}{%
\begin{tabular}{cccccccc} 

\toprule
Density  & TDE && Robustness  & TDE && \# Lanes  & TDE\\
\midrule
$N=5$ & $0.08$s && $\epsilon = 10^{-4}$ & $0.001$s && $L=2$ & $0.10$s\\
$N=13$ & $0.15$s && $\epsilon = 10^{-3}$ & $0.001$s && $L=4$ & $0.27$s\\
$N=20$ & $0.56$s && $\epsilon = 10^{-2}$ &$0.013$s && $L=6$ & $0.53$s\\
$N=25$ & $0.79$s && $\epsilon = 10^{-1}$ &$0.050$s && $L=8$ & $0.98$s\\
\bottomrule
\end{tabular}
}
\label{tab: sim_analysis}
\end{table}

One of the main issues with these datasets is that they do not contain labels for aggressive and conservative driving behaviors. Therefore, we obtain ground-truth driver behavior annotations using Algorithm~\ref{alg: ex}. We directly use the raw trajectory data from these datasets without any pre-processing or filtering step. For each video, the final ground-truth annotation (or label) is the expected value of the frame at which the ego-vehicle is most likely to be executing an aggressive driving style. This is denoted as $\EX[T]$. The goal for any driver behavior prediction model should be to predict the aggressive style at a time stamp as close to $\EX[T]$ as possible. The implied difference between the two time stamps is measured by the TDE metric.


The TDE metric is computed by Equation~\ref{eq: TDE}. Here, $t_\textrm{SLE} =  \argmax_{t \in \Delta t}{\textrm{SLE}(t)}$, as explained in Section~\ref{sec: approach}. We use Algorithm~\ref{alg: ex} for computing $\EX[T]$. We recruited $M=35$ participants with driving experience in at least two countries out of USA, Singapore, China, and India. This ensured that participants are ``expert'' annotators we are able to obtain gold-standard labels. For each video, every participant was asked to mark the starting and end frames for the time-period during which a vehicle is observed executing an aggressive maneuver. Participants were asked to watch out for typical traits such as overspeeding, overtaking, sudden lane-changes, weaving, driving slowly in single lanes etc. Once the start and end frames are recorded, we proceed by using Algorithm 1 as explained in Section V-B(b). Participants were allowed to scrub back and forth during a video and replay any moment any number of times. Furthermore, participants were allowed to zoom into a video to inspect styles more closely. We ignore repetitions and did not observe any control errors.

For each video, we end up with $S = \{s_1, s_2, \ldots, s_M\}$ and $E = \{e_1, e_2, \ldots, e_M\}$ start and end frames, respectively. We extract the overall start and end frame by finding the minimum and maximum value in $S$ and $E$, respectively (lines $1-2$). We denote these values as $s^*$ and $e^*$. Next, we initialize a distinct counter, $c_t$, for each frame $t \in [s^*, e^*]$ (line $3$). We increment a counter $c_t$ by $1$ if $t \in [s_m, e_m]$ (lines $4-7$). The value of the counter $c_t$ is assigned to $\mathcal{P}(T)$ (line $8$). The $\EX[T]$ of $\mathcal{P}(T)$ can then be computed using the standard definition of expectation of a discrete probability mass function (line $10$). Algorithm~\ref{alg: ex} is applied separately for each video in each dataset.







\subsection{Validating StylePredict Using TDE} 
\label{subsec: real_world_traffic_analysis}
\begin{figure*}[t]
\centering
\begin{subfigure}[h]{0.24\textwidth}
    \includegraphics[width=\textwidth]{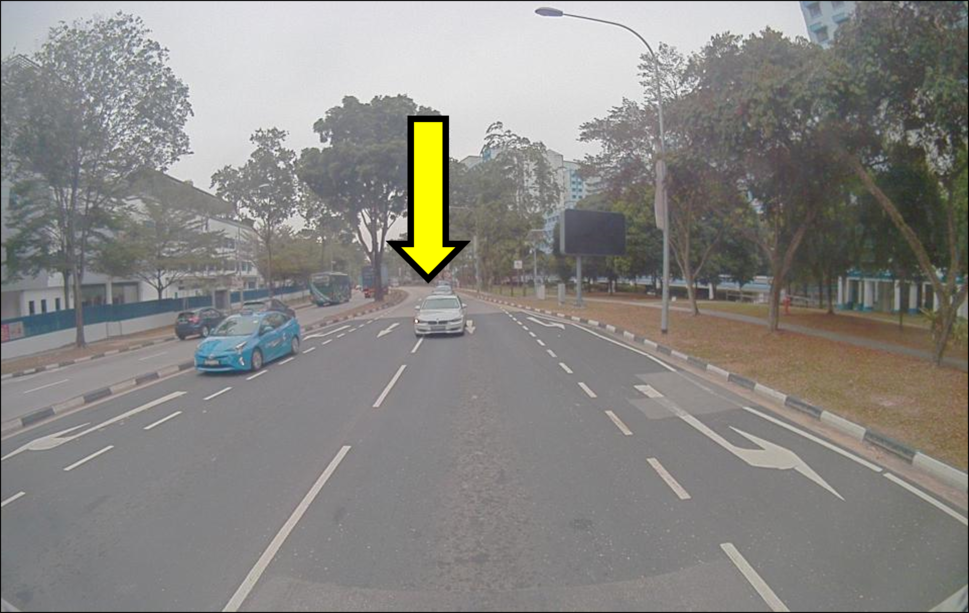}
    \caption{Frame $65$.}
    \label{fig: sg1}
  \end{subfigure}
   \begin{subfigure}[h]{0.24\textwidth}
    \includegraphics[width=\textwidth]{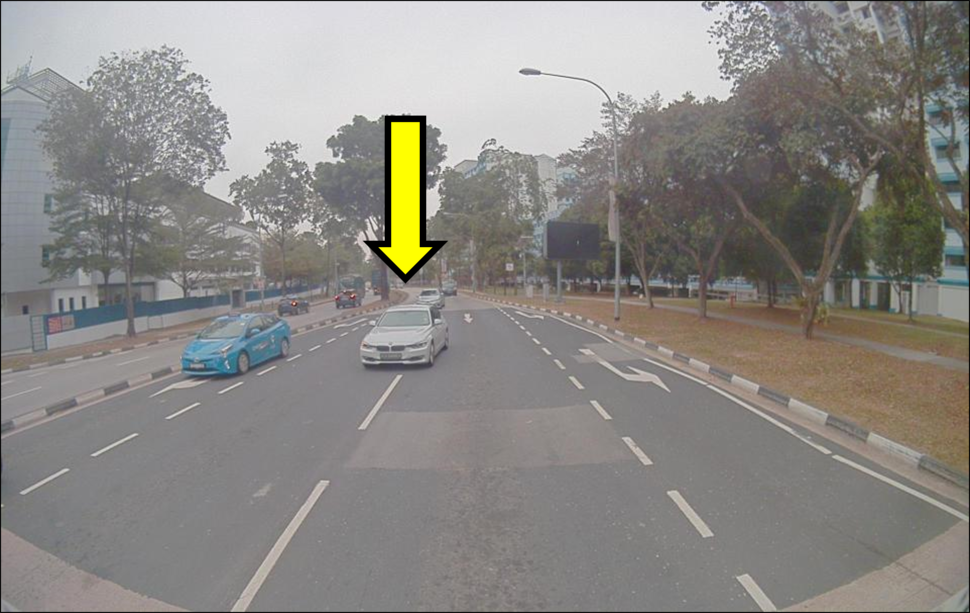}
    \caption{Frame $70$.}
    \label{fig: sg2}
  \end{subfigure}
  \begin{subfigure}[h]{0.24\textwidth}
    \includegraphics[width=\textwidth]{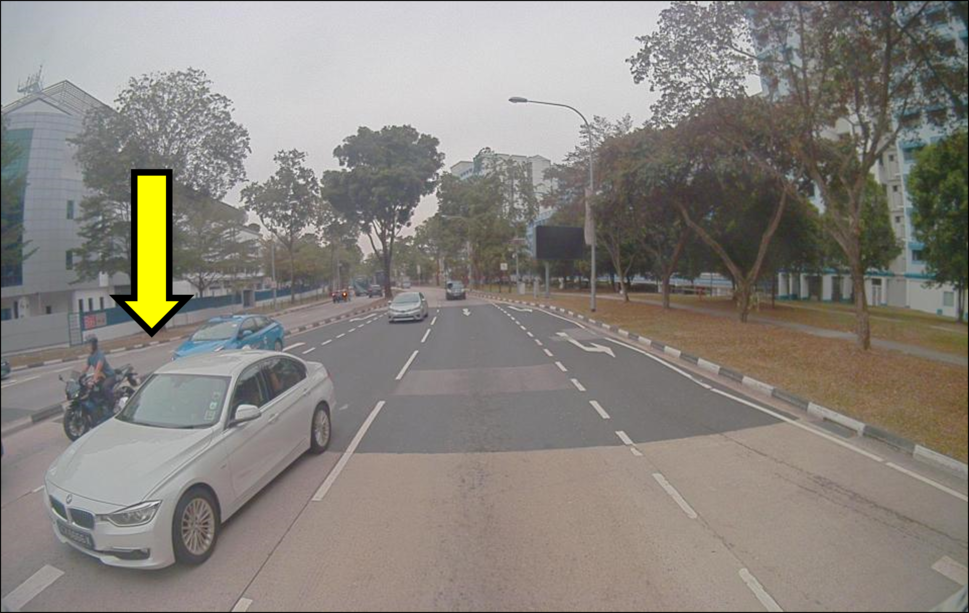}
    \caption{Frame $75$.}
    \label{fig: sg3}
  \end{subfigure}
  \begin{subfigure}[h]{0.24\textwidth}
    \includegraphics[width=\textwidth]{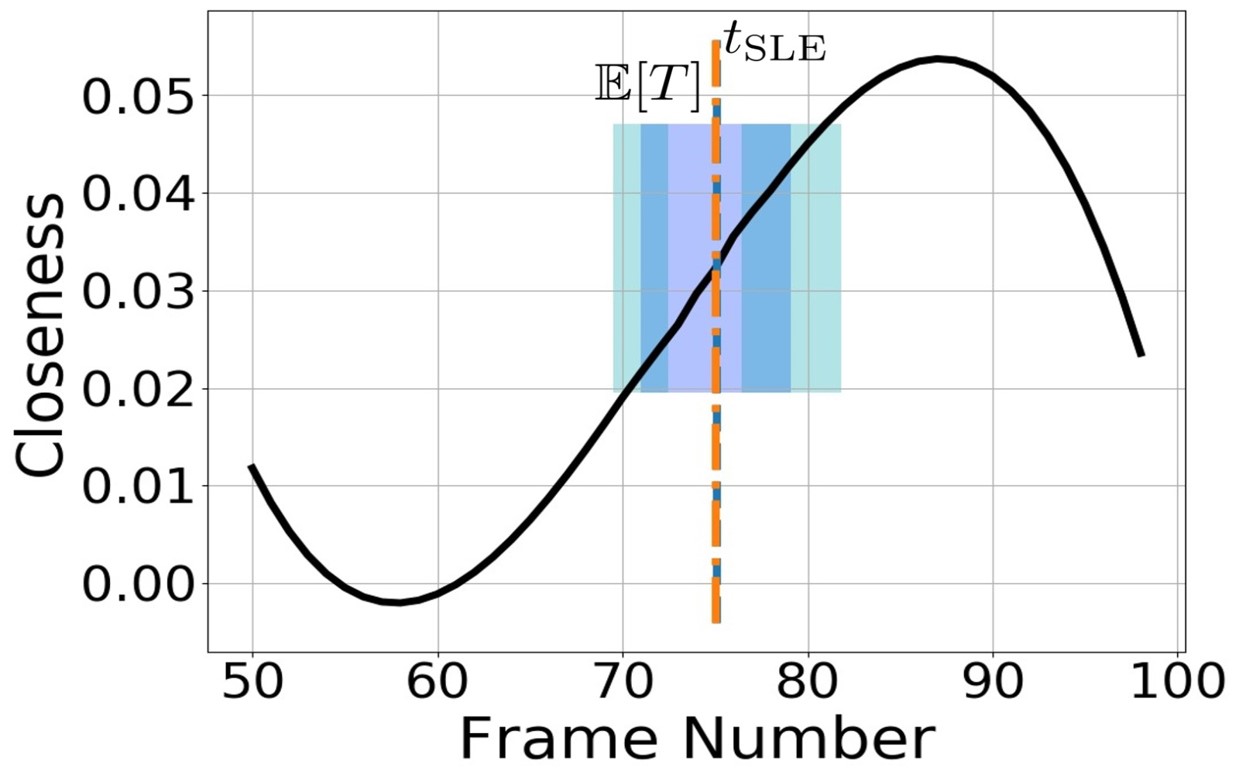}
    \caption{Sudden Lane-Change.}
    \label{fig: sg4}
  \end{subfigure}
  %
  %
    \begin{subfigure}[h]{0.24\textwidth}
    \includegraphics[width=\textwidth]{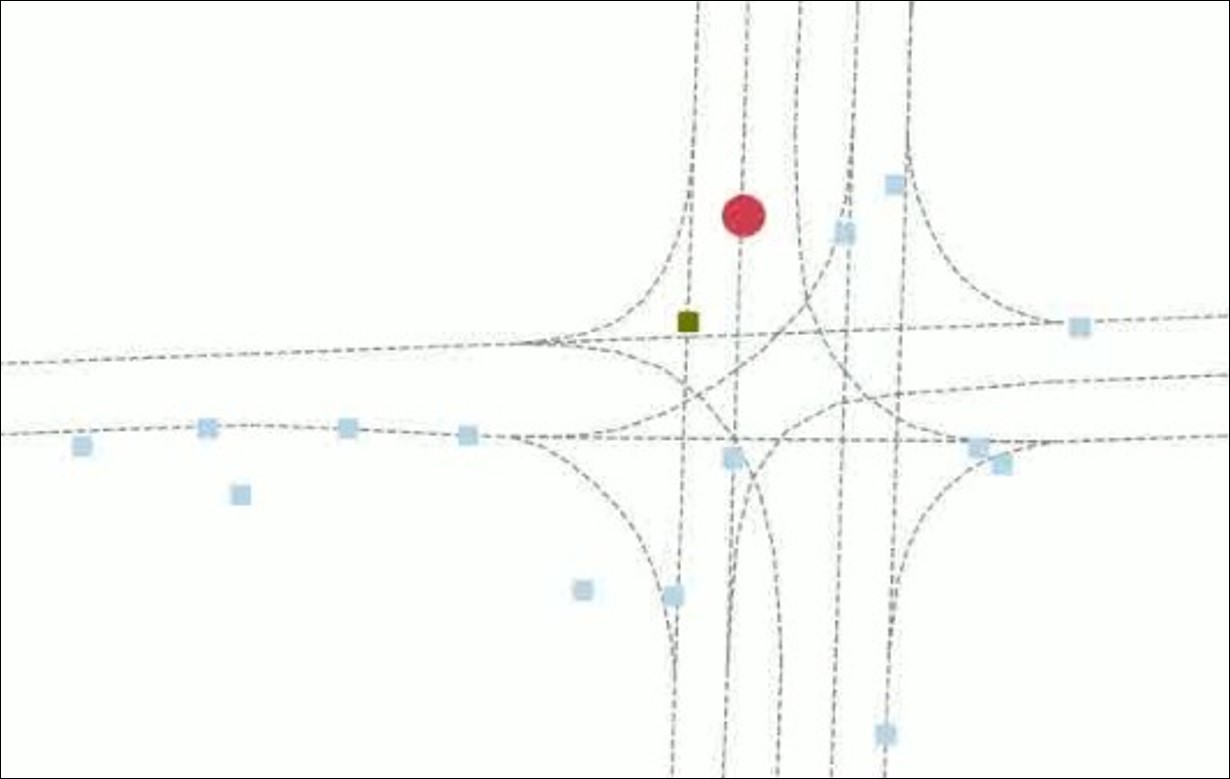}
    \caption{Frame $20$.}
    \label{fig: argo1}
  \end{subfigure}
  \begin{subfigure}[h]{0.24\textwidth}
    \includegraphics[width=\textwidth]{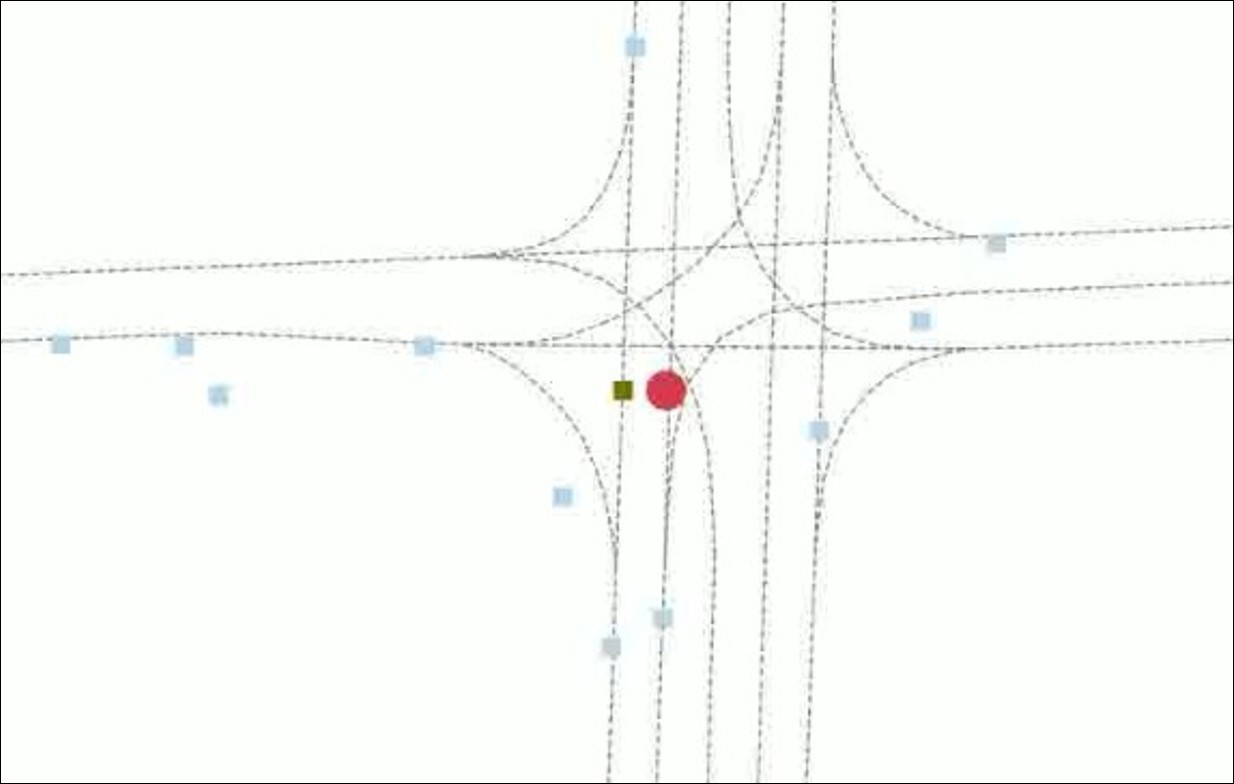}
    \caption{Frame $25$.}
    \label{fig: argo2}
  \end{subfigure}
    \begin{subfigure}[h]{0.24\textwidth}
    \includegraphics[width=\textwidth]{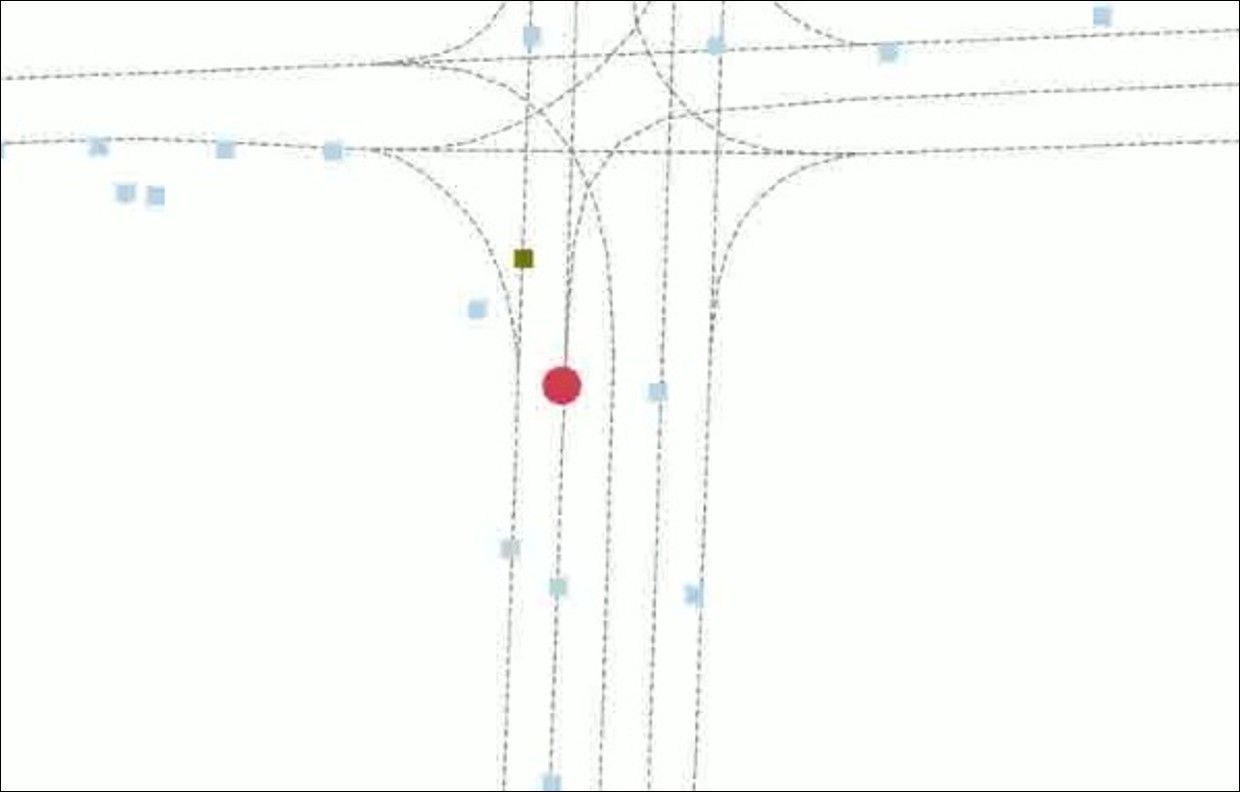}
    \caption{Frame $20$.}
    \label{fig: argo3}
  \end{subfigure}
    \begin{subfigure}[h]{0.24\textwidth}
    \includegraphics[width=\textwidth]{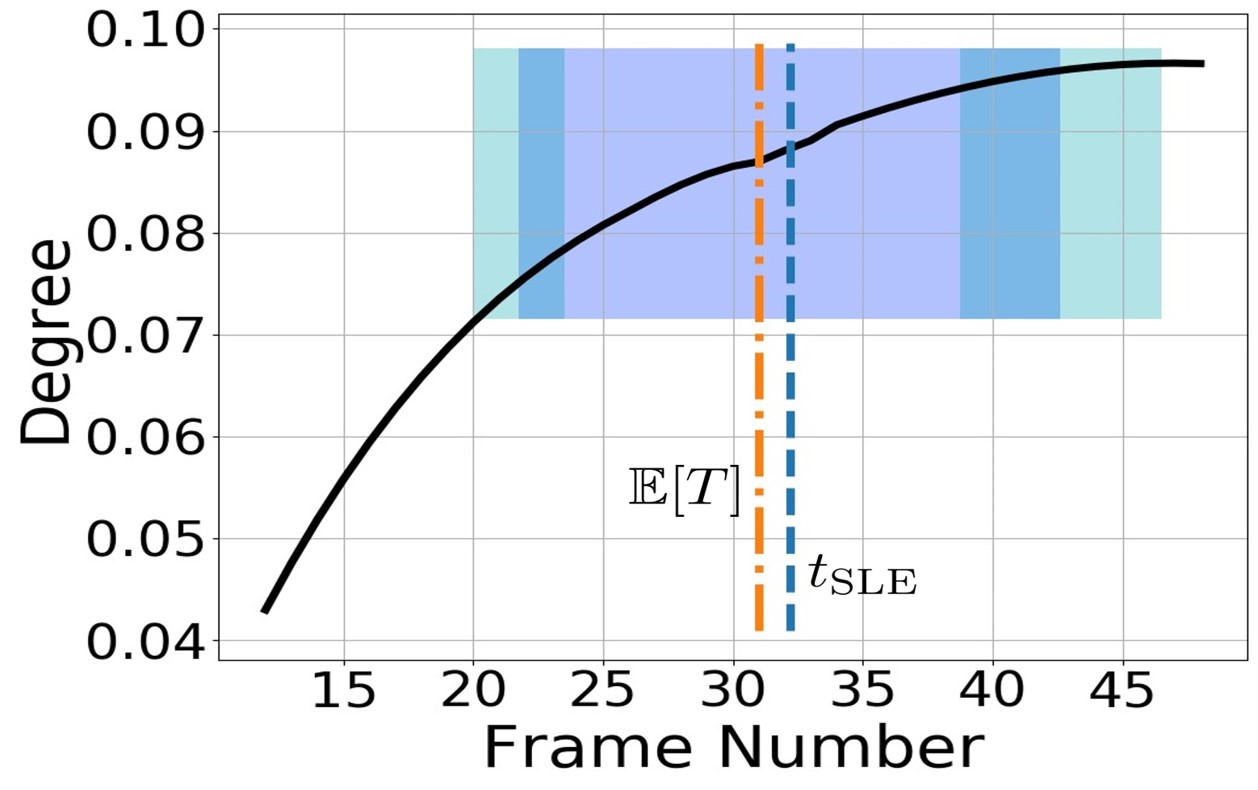}
    \caption{Overspeeding.}
    \label{fig: argo4}
  \end{subfigure}
    \begin{subfigure}[h]{0.24\textwidth}
    \includegraphics[width=\textwidth]{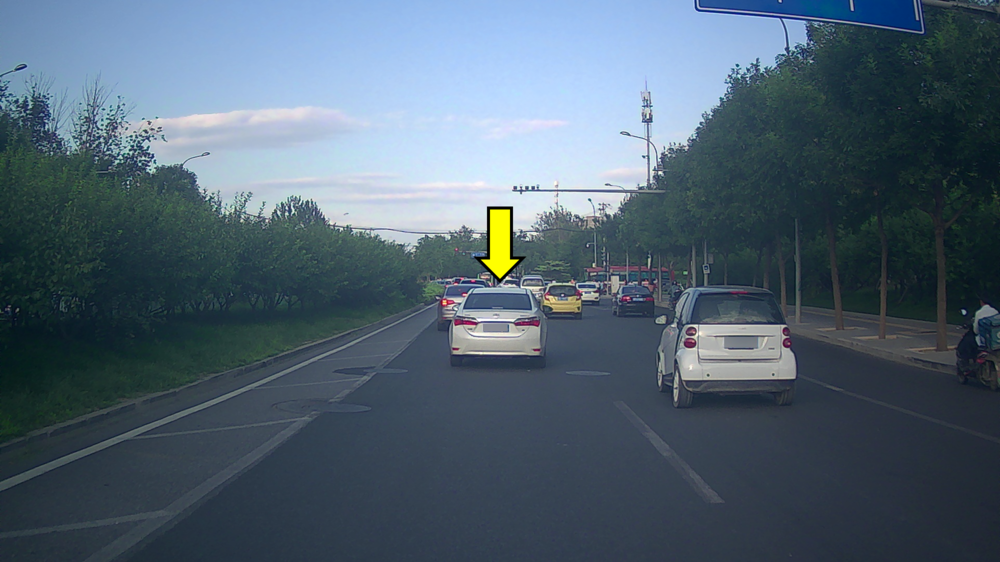}
    \caption{Frame $75$.}
    \label{fig: argo1}
  \end{subfigure}
  \begin{subfigure}[h]{0.24\textwidth}
    \includegraphics[width=\textwidth]{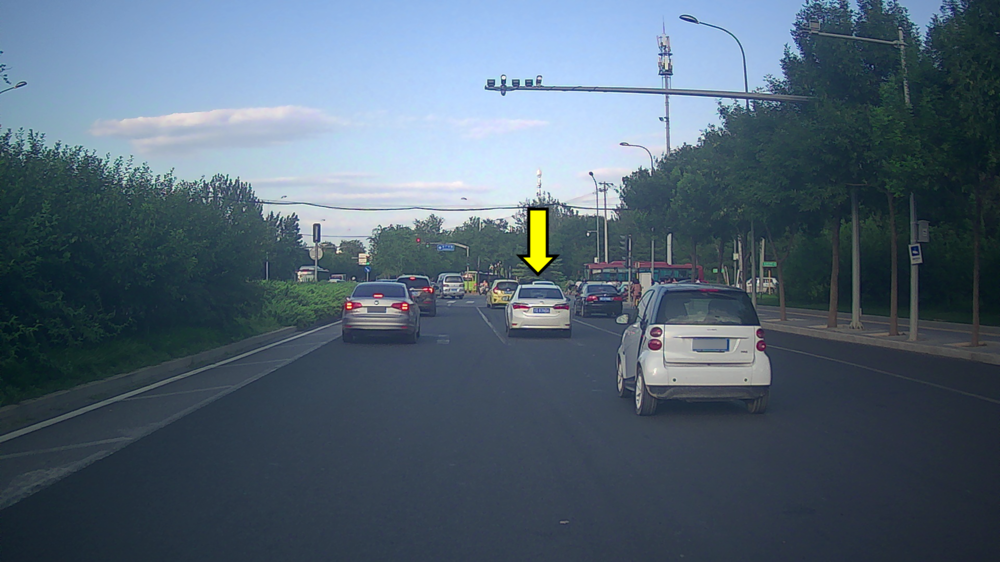}
    \caption{Frame $80$.}
    \label{fig: argo2}
  \end{subfigure}
    \begin{subfigure}[h]{0.24\textwidth}
    \includegraphics[width=\textwidth]{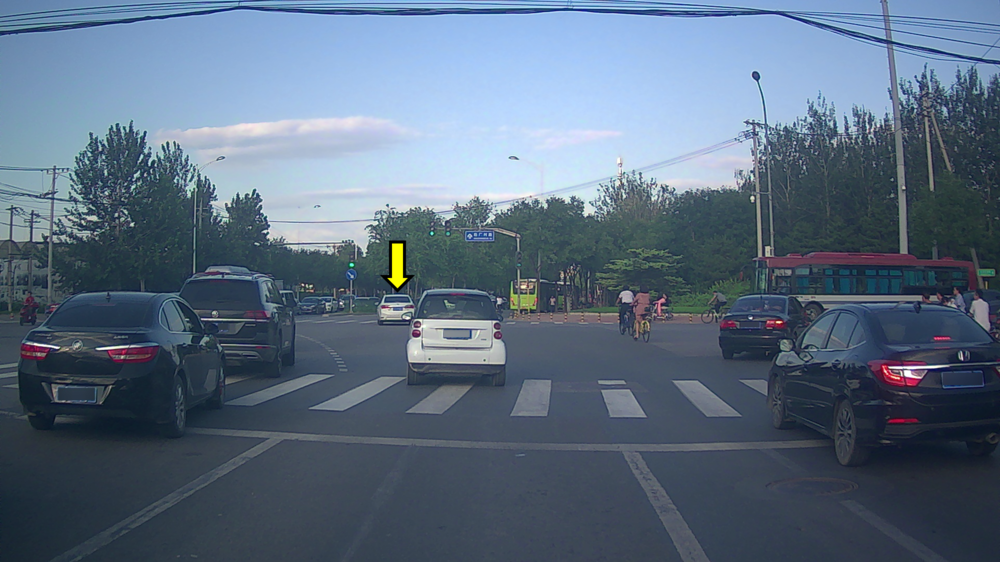}
    \caption{Frame $85$.}
    \label{fig: argo3}
  \end{subfigure}
    \begin{subfigure}[h]{0.24\textwidth}
    \includegraphics[width=\textwidth]{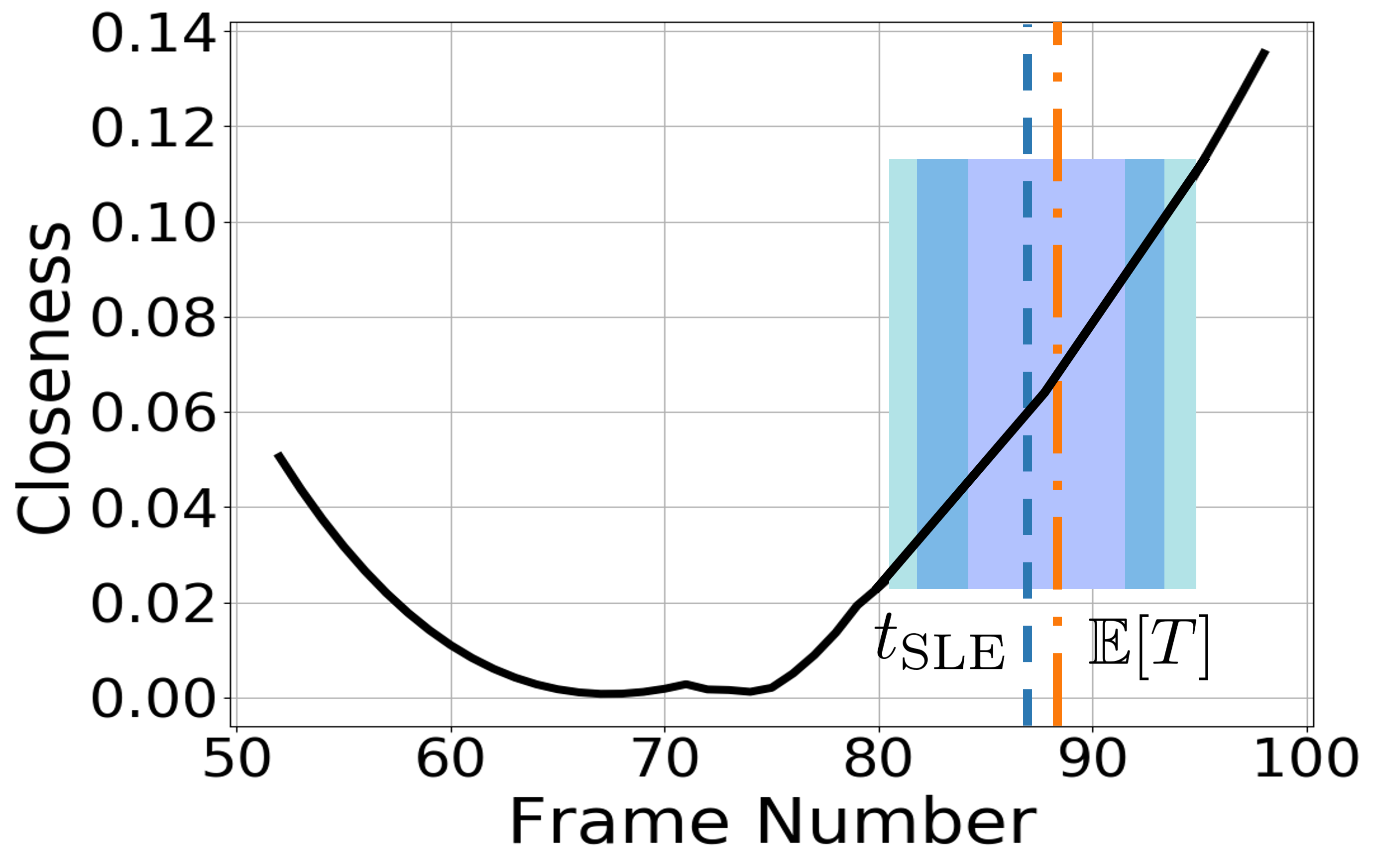}
    \caption{Weaving.}
    \label{fig: argo4}
  \end{subfigure}
    \begin{subfigure}[h]{0.24\textwidth}
    \includegraphics[width=\textwidth]{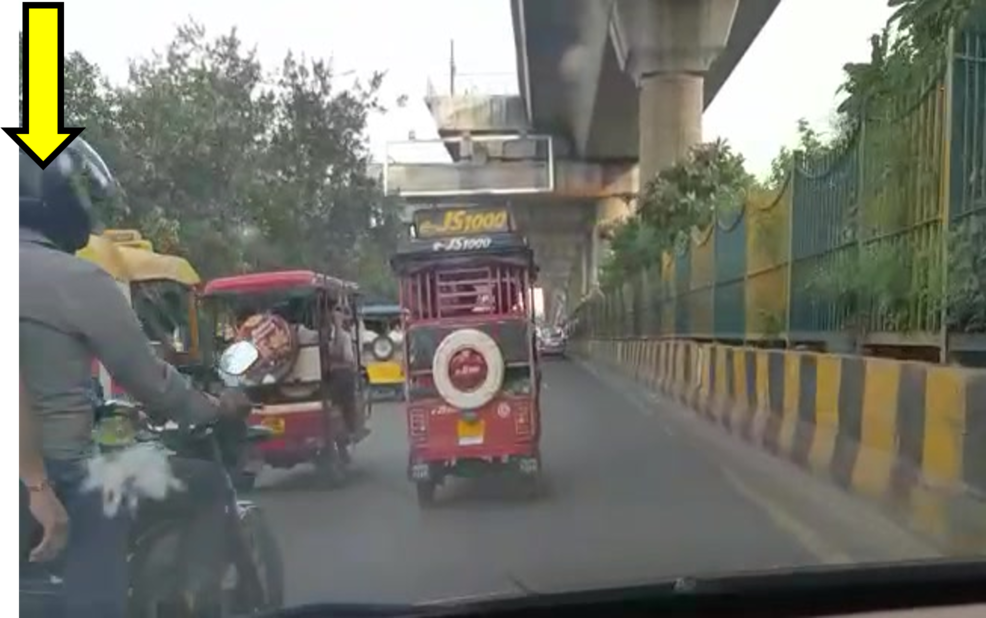}
    \caption{Frame $66$.}
    \label{fig: argo1}
  \end{subfigure}
  \begin{subfigure}[h]{0.24\textwidth}
    \includegraphics[width=\textwidth]{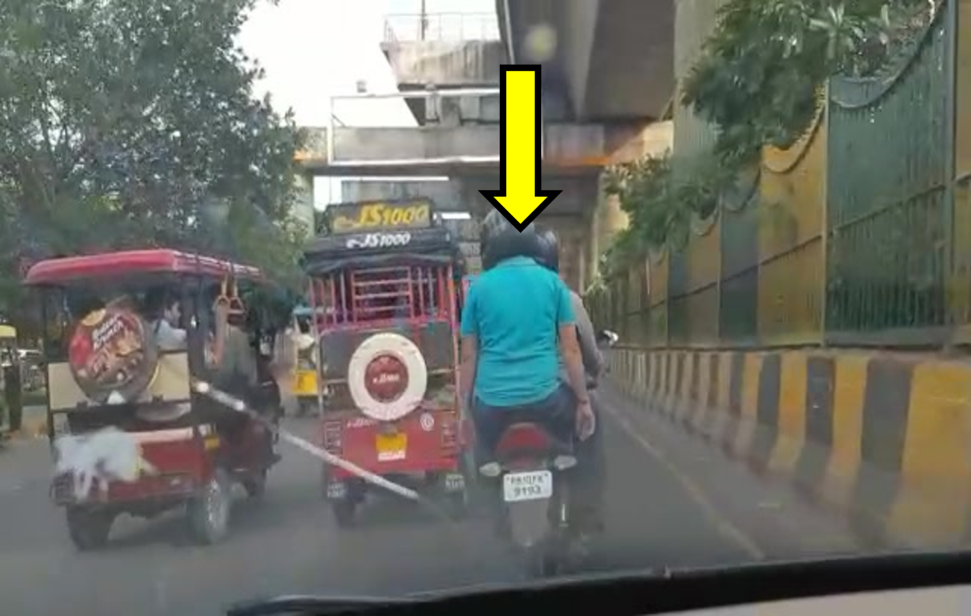}
    \caption{Frame $68$.}
    \label{fig: argo2}
  \end{subfigure}
    \begin{subfigure}[h]{0.24\textwidth}
    \includegraphics[width=\textwidth]{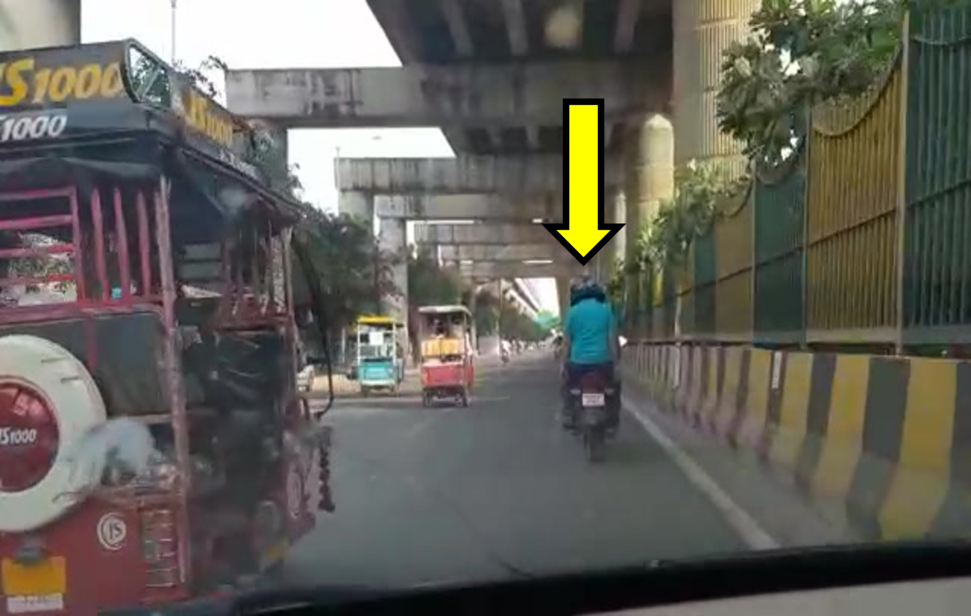}
    \caption{Frame $70$.}
    \label{fig: argo3}
  \end{subfigure}
    \begin{subfigure}[h]{0.24\textwidth}
    \includegraphics[width=\textwidth]{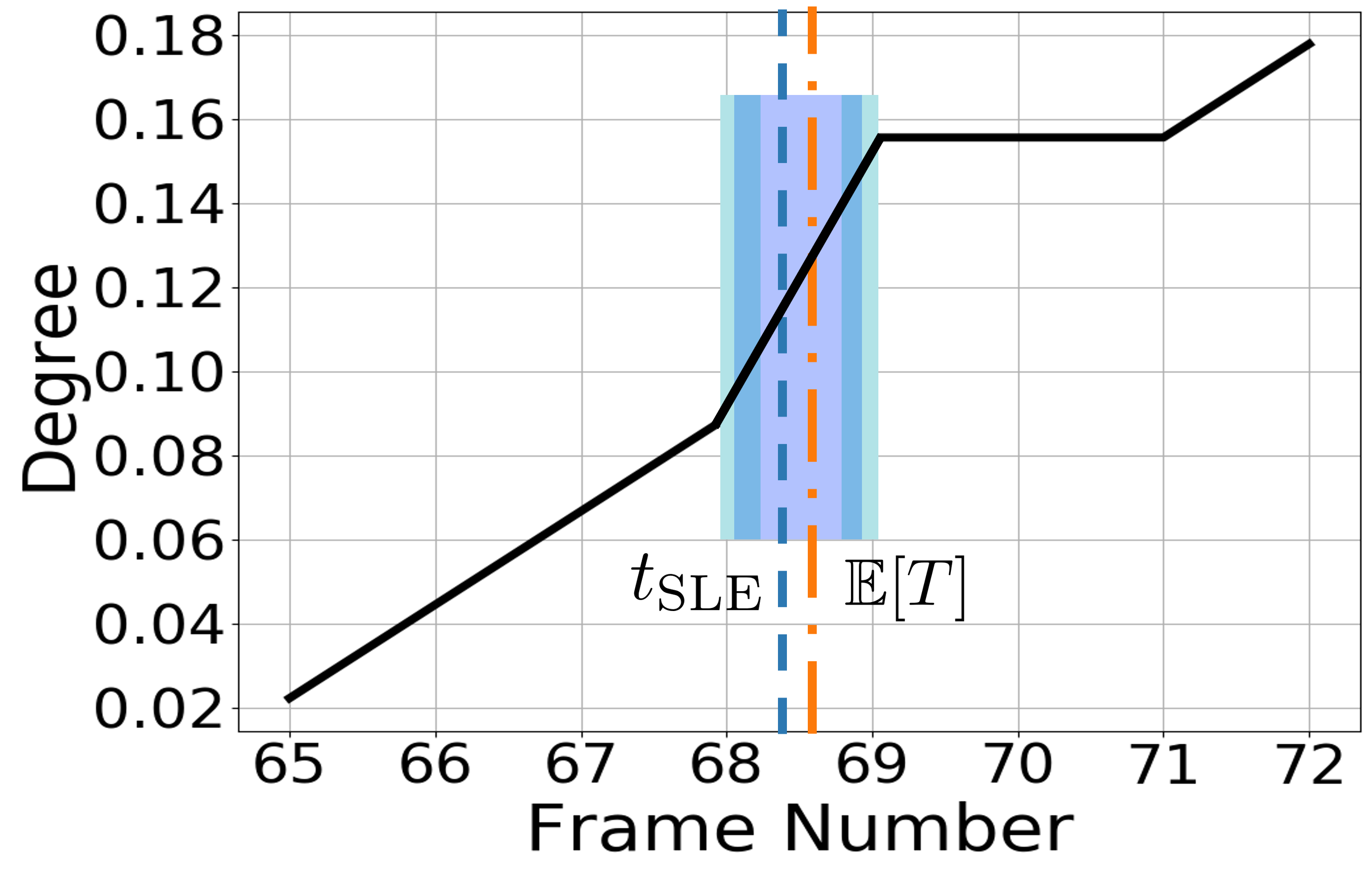}
    \caption{Overspeeding.}
    \label{fig: argo4}
  \end{subfigure}
\caption{\textbf{Driver Behavior Modeling in Singapore \textit{(top row)}, U.S. \textit{(second row)}, China \textit{(third row)}, and India \textit{(bottom row)}:} In each row, the first three figures demonstrate the trajectory of a vehicle executing an aggressive driving style (sudden lane change, overspeeding, weaving, and overspeeding, respectively), while the fourth figure shows the corresponding closeness or degree centrality plot. The shaded colored regions overlaid on the graphs in the first two rows are color heat maps that correspond to $\mathcal{P}(T)$ (line $8$, Algorithm~\ref{alg: ex}).}
  \label{fig: qualitative}
\end{figure*}
\begin{table}[t]
\centering
\caption{We report the Time Deviation Error (TDE) (in seconds (s)) for the following driving styles: Overspeeding (OS), Overtaking (OT), Sudden Lane-Changes (SLC), and Weaving (W) along with their \% appearance in various real-world datasets. On average, we find that it is easiest to predict weaving and sudden lane-changes in India. This observation agrees with our cultural analysis in Section~\ref{subsec: real_world_traffic_analysis}}

\centering
\resizebox{\columnwidth}{!}{%
\begin{tabular}{lcccccccc} 
\toprule
\multirow{3}{*}{Dataset}  &  \multicolumn{8}{c}{Styles} \\
& \multicolumn{2}{c}{OS}  & \multicolumn{2}{c}{OT} & \multicolumn{2}{c}{SLC} & \multicolumn{2}{c}{W}\\
\cmidrule{2-9}
& TDE & \%& TDE & \%& TDE & \%& TDE & \% \\
\midrule
U.S.~\cite{Argoverse}  & $0.25$s& $83$ & $0.67$s& $2$ & $0.23$s& $14$& $0.26$s&$1$  \\
Singapore & $0.54$s& $27$ & $0.88$s& $27$ & $1.21$s & $27$ & $1.28$s& $18$ \\
China~\cite{wang2019apolloscape} & $0.74s$& $24$ & $0.44$s& $32$ & $0.39$s& $36$ & $0.23$s& $8$\\
India~\cite{chandra2019traphic}  & $0.81$s& $16$ & $0.38$s& $40$ & $0.19$s& $28$& $0.06$s& $16$\\
\bottomrule
\end{tabular}
}
\label{tab: accuracy}
%
\end{table}

In Table~{\ref{tab: sim_analysis}}, we report the average TDE in seconds (s) in simulation environments. We used Algorithm~{\ref{alg: ex}} to compute the TDE in various simulation settings. First, we varied the traffic density by increasing the number of vehicles from $5$ to $25$. As the number of vehicles grows, the TDE increases, which is to be expected since it is harder for a human participant to spot different styles in denser traffic resulting in a detection delay and therefore higher TDE. Next, we analyzed the robustness property by varying the noise parameter $\epsilon$ (Equation~{\ref{eq: noisy_OLS}}). We opted for $\epsilon = \{ 10^{-4},10^{-3},10^{-2},10^{-1} \}$ as this range reflects the most common values of error that may occur in nature. TDE for values lower than $10^{-4}$ all converged to $0$. We naturally observe that the TDE increases with more noise. Finally, we varied the number of lanes from $2-8$ ($1$ lane is invalid as lateral styles cannot be observed in a single lane) and observe that the TDE increases with the number of lanes. This is because, with more space available, it is unclear to human participants whether a particular maneuver is aggressive or neutral. On the other hand, overtaking and lane-changing in a $2-$lane highway is very evident and easy to spot, resulting in a lower TDE.

In Table~\ref{tab: accuracy}, we report the average TDE in seconds (s) in different geographical regions and cultures for the following driving styles: Overspeeding (OS), Overtaking (OT), Sudden Lane-Changes (SLC), and Weaving (W). 
The traffic conditions differ significantly due to the varying cultural norms in different countries such as Singapore, the United States (U.S.), China, and India. For instance, traffic is more regulated in the U.S. than in Asian countries such as India or China, where vehicles do not conform to standard rules such as lane-driving. Such differences contribute to different driving behaviors. Our quantitative results in Table~\ref{tab: accuracy} and qualitative results in Figure~\ref{fig: qualitative} show that our driver behavior modeling algorithm is not affected by cultural norms. Across all cultures, the average TDE is less than $1$ second for every specific style. Aggressive vehicles are still associated with high centrality values, while conservative vehicles remain associated with low centrality values.

In Figure~\ref{fig: qualitative}, we show traffic recorded in Singapore \textit{(top row)}, the U.S. \textit{(second row)}, China \textit{(third row)}, and India \textit{(bottom row)}. In each scenario, the first three columns depict the trajectory of a vehicle executing a specific style between some time intervals. The last column shows the corresponding centrality plot. The shaded colored regions overlaid on the graphs in Figures~\ref{fig: sg4} and~\ref{fig: argo4} are color heat maps that correspond to $\mathcal{P}(T)$ (line $8$, Algorithm~\ref{alg: ex}). The orange dashed line indicates the mean time frame, $\EX[T]$, and the blue dashed line indicates $t_\textrm{SLE}$. The main result can be observed by noting the negligible distance between the two dashed lines, \textit{i.e.} the TDE.

In the first row (corresponding to traffic in Singapore), for instance, our approach accurately predicts a maximum likelihood of a sudden lane-change by the white sedan at around the $75^\textrm{th}$ frame (blue dashed line, Figure~\ref{fig: sg4}), with an average TDE of $0.88$ seconds. Similarly, in the second row (corresponding to traffic in the U.S.), we precisely predict the maximum likelihood of the vehicle overspeeding by the vehicle denoted by the red dot at around the $30^\textrm{th}$ frame with a TDE of $0.25$ seconds. Note that in both cases the TDE (the distance between the blue and orange dashed vertical lines) is $<1$ second. 

\subsection{Analyzing Behavior Prediction Using Weighted Accuracy}
\label{subsec: behaviore_prediction_eval}

We compare our approach with Dynamic Attributed Network Embedding (DANE)~\cite{dane} and Cheung et al.~\cite{ernest}. Both baselines predict human behavior but differ in their techniques. DANE also uses a graph-based approach (although not based on centrality) where the main step consists of computing the spectrum of the Laplacian matrix. Cheung et al., on the other hand, use linear lasso regression on trajectory features, extracted from raw traffic videos. We present a comparison using the weighted accuracy with DANE and Cheung et al. in Table~\ref{tab: accuracy_ML} where we show an improvement of up to~$25$\%. 

\begin{figure*}[t]
\centering
\begin{subfigure}[h]{0.32\linewidth}
    \includegraphics[width=\textwidth]{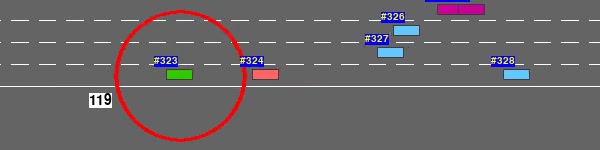}
    \caption{AV approaches aggressive vehicle.}
    \label{fig: slow_down_1}
  \end{subfigure}
   \begin{subfigure}[h]{0.32\linewidth}
    \includegraphics[width=\textwidth]{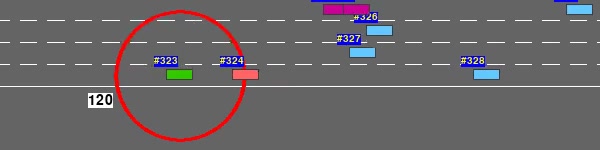}
    \caption{AV slows down, but does not overtake .}
    \label{fig: slow_down_2}
  \end{subfigure}
  \begin{subfigure}[h]{0.32\linewidth}
    \includegraphics[width=\textwidth]{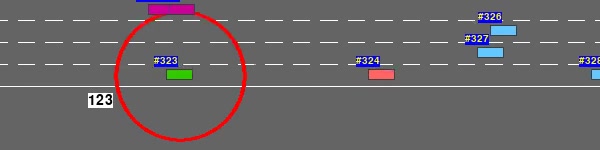}
    \caption{AV resumes acceleration.}
    \label{fig: slow_down_3}
  \end{subfigure}
  \begin{subfigure}[h]{0.32\linewidth}
    \includegraphics[width=\textwidth]{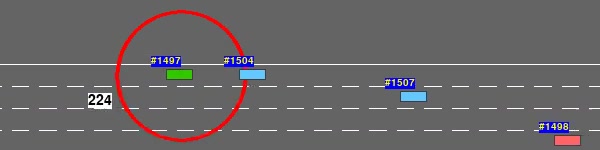}
    \caption{AV approaches conservative vehicle.}
    \label{fig: lane_change_1}
  \end{subfigure}
    \begin{subfigure}[h]{0.32\linewidth}
    \includegraphics[width=\textwidth]{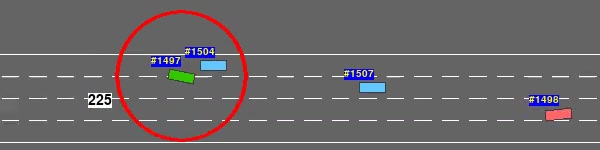}
    \caption{AV confidently switches to adjacent lane.}
    \label{fig: lane_change_2}
  \end{subfigure}
  \begin{subfigure}[h]{0.32\linewidth}
    \includegraphics[width=\textwidth]{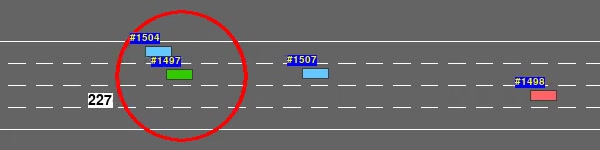}
    \caption{AV overtakes conservative vehicle.}
    \label{fig: lane_change_3}
  \end{subfigure}
\caption{\textbf{Behavior-Guided Navigation:} While interacting with aggressive (\textit{top}) and conservative (\textit{bottom}) vehicles. We indicate the AV, aggressive, and conservative vehicles in green, red, and blue, respectively. (\textit{Top}) The AV senses that the red vehicle is aggressive and therefore decides to slow down instead of overtaking. (\textit{Bottom}) The AV notices the conservative vehicle in front and decides to confidently overtake it. }
\label{fig: behavior_navigation}
\end{figure*}

\begin{table}[t]
\caption{We compare the weighted classification accuracy of StylePredict versus supervised learning-based SOTA methods on the Argoverse dataset~\cite{Argoverse}. Additionally, we compare the accuracy of different supervised learning machine learning and deep learning algorithms.}
\centering
\resizebox{\columnwidth}{!}{%
\begin{tabular}{ccc} 
\toprule
Dataset \Tstrut & Method \Bstrut &   Weighted Accuracy \\

\midrule
\multirow{6}{*}{Argoverse }
& DANE~\cite{dane} \Tstrut      & $65.50$\%\\
& Cheung et al.~\cite{ernest}     & $62.50$\%\\
& StylePredict \textit{w. LR} & $69.90$\% \\
& StylePredict \textit{w. RNN} & $70.80$\% \\
& StylePredict \textit{w. SVM} & $75.00$\% \\
& \textbf{StylePredict \textit{w. MLP}}              & $\textbf{89.90\%}$ \\
\bottomrule
\end{tabular}
}
\label{tab: accuracy_ML}
\vspace{-15pt}
\end{table}

StylePredict uses a multi-layer perceptron (MLP)~\cite{mlp} for the classification task. However, other classifiers in the machine learning literature such as logistic regression (LR), support vector machines (SVM)), and deep neural networks (RNNs) may be used. In Table~\ref{tab: accuracy_ML}, we compare the results of replacing the MLP with different classifiers and benchmark the different output accuracies against the MLP.
\section{Behaviorally-Guided Navigation}
\label{sec: bg_navigation}

In autonomous navigation, it is important to handle the unpredictable and/or aggressive nature of human drivers (See Section~\ref{sec: introduction}). In this section, we show that, unlike existing navigation methods~\cite{schwarting2018planning}, StylePredict can be used to train a behaviorally-guided navigation policy that takes into account the conservative or aggressive nature of human drivers. We begin by generating behaviorally-guided trajectories for vehicles in existing traffic simulators~\cite{leurent2019social} using StylePredict for training a reinforcement learning (RL) navigation policy in Section~\ref{subsec: sim_traj}. This is followed by a discussion of the RL model used to obtain the behaviorally-guided navigation policy in Section~\ref{subsec: RL_policy}. Finally, in Section~\ref{subsec: analyze_nav}, we show that such a behaviorally-guided navigation policy allows an AV to perform more efficient lane changes and adapt its speed according to the behavior of the vehicles around it.

\subsection{Augmenting Traffic Simulators with Driver Behavior}
\label{subsec: sim_traj}

Existing traffic simulators~\cite{leurent2019social, sumo} assume a fixed motion model that does not take into account the nature of other drivers. Such simulators inevitably produce navigation policies that are not behaviorally-guided. Therefore, we use the StylePredict algorithm to augment such traffic simulators with driver behavior information such that they generate behaviorally-guided trajectories. The behavior of a vehicle in the Highway-Env simulator~\cite{leurent2019social} is controlled using a set of parameters described in the supplementary material. To simulate aggressive and conservative behaviors, we find the appropriate range for these parameters by iteratively performing the following steps:

\begin{enumerate}
    \item We initialize the simulator parameters as random values.
    
    \item We use the parameters to simulate the trajectories for each vehicle. By varying the parameters, the vehicles perform longitudinal and lateral maneuvers with certain likelihoods and intensities.
    
    \item We use StylePredict to measure these likelihoods (SLE) and intensities (SIE) following the framework introduced in Section~\ref{sec: stylepredict}. These measures indicate the aggressiveness or conservativeness of a driver. For instance, a high likelihood and intensity of lane changing and overspeeding would indicate an aggressive driver. Based on this feedback, we update the parameters for the next episode.
    
    \item We repeat steps $2$ and $3$ until the likelihoods and intensities of maneuvers match those of a desired behavior (aggressive or conservative).
    
\end{enumerate}

\noindent The final set of parameter values corresponding to aggressive and conservative behaviors is used to instantiate aggressive and conservative vehicles at runtime, shown as red and blue vehicles in Figure~\ref{fig: qualitative}. Our enhanced traffic simulator results in trajectories with varying levels of aggressiveness, in terms of maneuvers like overspeeding, overtaking, and so on.

    
    
    
    


\subsection{ Training the Behaviorally-Guided Navigation Policy}
\label{subsec: RL_policy}

We use RL to learn a behaviorally-guided navigation policy offline using the enhanced simulator with behaviorally-augmented trajectories, and deploy it at runtime. We formulate the RL model as a Markov Decision Process (MDP) and use deep Q-learning~\cite{dqn} to train the navigation policy. We refer the reader to~\cite{bgap} (Sections IV and V) for further details on the RL model.

We frame the navigation problem as a Markov Decision Process (MDP) represented by $\mathcal{M} := \{\mathcal{S},\mathcal{A},\mathcal{T},\gamma,\mathcal{R}\}$. The sets of possible states and actions are denoted by $\mathcal{S}$ and $\mathcal{A}$, respectively. $\mathcal{T}: \mathcal{S} \times \mathcal{S} \times \mathcal{A} \rightarrow \mathbb{R}$ captures the state transition dynamics,  $\gamma$ is the discounting factor, and $\mathcal{R}$ is the set of rewards defined for all of the states in the environment. The state of the world $S$ at any time step is equal to a matrix $F\times V$, which includes the state $s$ of every vehicle in the environment. $V$ is the number of vehicles considered, and $F$ is the number of features used to represent the state of a vehicle. The environment consists of a highway road with four single-direction lanes, along with conservative and aggressive traffic agents that are generated using StylePredict. The ego-vehicle can take five different actions: $\mathcal{A}=$\{"accelerate", "decelerate", "right lane-change", "left lane-change", "idle"\}. The motivation behind the reward function $\mathcal{R}$ is based on our objective for training an agent that can safely and efficiently navigate in dense traffic while respecting other road agents in its neighborhood.  The state transition matrix $\mathcal{T}$ boils down to a state transition probability $P(s^\prime|s)$, which is defined as the probability of beginning from a current state $s$ and arriving at a new state $s^\prime$. This probability is calculated by the kinematics of the simulator, which depends on the underlying motion models, and thus it is equal to $1$, establishing a deterministic setting.

\begin{table}[t]
\caption{ We measure the average speed (Avg. Spd.) and the number of lane changes (\#LC) for the ego-vehicle when it is (left) navigating in the default simulator (without any varying behaviors), (middle) interacting with conservative agents only, and (right) interacting with a combination of conservative and aggressive traffic-agents. }
\centering
\resizebox{\columnwidth}{!}{%
\begin{tabular}{ccccccc} 
\toprule
Model& \multicolumn{2}{c}{Default}& \multicolumn{2}{c}{Conservative} & \multicolumn{2}{c}{Aggressive} \\
 ($n=20$)&Avg. Spd. (m/s) & \#LC & Avg. Spd. (m/s) & \#LC & Avg. Spd. (m/s) & \#LC  \\
\midrule
MLP & $22.65$ & $4.1$ & $19.7$ & $2.9$ & $28.8$ & $2.6$\\
GCN & $22.26$ & $0.82$ & $18.9$ & $2.33$ & $29$ & $1.4$\\
\midrule
Model& \multicolumn{2}{c}{Default}& \multicolumn{2}{c}{Conservative} & \multicolumn{2}{c}{Aggressive} \\
 ($n=10$)&Avg. Spd. (m/s) & \#LC & Avg. Spd. (m/s) & \#LC & Avg. Spd. (m/s) & \#LC  \\
\midrule
MLP & $23.75$ & $6.25$ & $21.4$ & $3.5$ & $29.16$ & $2.06$\\
GCN & $23.6$ & $0.35$ & $20.6$ & $1.6$ & $28.9$ & $1.3$\\
\bottomrule
\end{tabular}
}
\label{tab: evaluation}
\vspace{-10pt}
\end{table}

\subsection{Benefits of Behaviorally-Guided Navigation}
\label{subsec: analyze_nav}

\noindent \textbf{Experiment Setup:} 
We evaluate both dense ($N=20$) and sparse ($N=10$) highway traffic scenarios, where $N$ represents the number of vehicles. We perform experiments with two different RL policies using a graph convolutional network (GCN)~\cite{gcn_review} and a multi-layer perceptron (MLP)~\cite{mlp}. MLP and GCN are two different neural network architectures used to train a navigation policy using deep reinforcement learning. The choice of the underlying architecture determines, in part, the type of navigation policy trained and may result in different characteristics. 
The MLP and the GCN differ in several ways, including input data format, applications, and internal operations. We employ Q-Learning~\cite{dqn} to learn an autonomous navigation policy. Specifically, we use the MLP and GCN to receive observations of the state space as input and implicitly model the behavioral interactions between aggressive and conservative agents and the ego-vehicle. Ultimately, the MLP and GCN learn a function that receives a feature matrix that describes the current state of the traffic as input and provides us with the optimal $Q$ values of the state space. Finally, the ego-vehicle can use the learned model during evaluation time in order to navigate its way around the traffic by choosing the best action that corresponds to the maximum $Q$ value for its state at every time step.

We apply two metrics to evaluate over $100$ episodes and average them at test time:
\begin{itemize}
    \item Average Speed of the AV, which captures the distance per second covered in a varying time interval.
    \item Number of lane changes performed by the AV on average during the given duration. In general, fewer lane changes imply that the ego vehicle can cover the same distance with fewer maneuvers. 
\end{itemize} 

\noindent Finally, we vary the behavior of traffic agents between conservative and aggressive behaviors and evaluate the performance of our action prediction and navigation algorithms. In the conservative environment (Table~\ref{tab: evaluation}, ``Conservative'' column), the environment is populated by conservative agents only, while in the aggressive environment (``Aggressive'' column), a significant number of the agents are aggressive, but there are also some conservative agents.



\noindent \textbf{Analysis and Insights:} We observe two advantages of behaviorally-guided navigation, which we discuss below:

\subsubsection{Intelligent Lane-Changing}
We observe that behavior-guided navigation allows AVs to switch lanes more intelligently, than those navigating without any knowledge of driver behavior. More specifically, in conservative traffic the AV learns to confidently overtake slow-moving traffic whereas in aggressive traffic the AV executes fewer lane changes around aggressive agents in its vicinity.

To support this claim, we visualize this result in Figure~\ref{fig: behavior_navigation}. In Figures~\ref{fig: slow_down_1}, \ref{fig: slow_down_2}, and \ref{fig: slow_down_3}, we show that the AV quickly approaches the aggressive vehicle from the rear. However, rather than overtaking it (the AV has sufficient space), the AV chooses to decelerate and wait to resume acceleration when it is safe to do so. In contrast, in Figures~\ref{fig: lane_change_1}, \ref{fig: lane_change_2}, and \ref{fig: lane_change_3}, we show that the AV (green) confidently overtakes the conservative vehicle (blue).

However, when navigating without driver behavior modeling, the AV does not follow a consistent lane-changing pattern. For instance, we see in Table~{\ref{tab: evaluation}} under the ``Default, $\#$LC'' column, that the AV performs an excessive number of lane changes with one policy (MLP), but mostly follows the traffic with very few lane changes with the other policy (GCN). Dependence of lane changes on the type of navigation policies is undesirable as the choice of policy often depends on several factors such as computational resources, type of application, different performance measuring metrics, and so on. These factors may vary for different situations and may potentially result in unsafe lane-changing in dense traffic. With the default simulator, the \# of lane changes performed by a vehicle depends \textit{exclusively} on the type of navigation policy (Table~{\ref{tab: evaluation}} - $4.1$ for MLP and $0.82$ for GCN in the dense traffic setting).


\subsubsection{Adapting Speed to Different Behaviors}
Behavior-guided navigation allows AVs to adapt their speeds to the nature of traffic around them. To be specific, AVs slow down more often in preparation to overtake slow moving conservative agents whereas they are able to maintain high speeds in fast moving aggressive traffic. We provide evidence for this result in Table~\ref{tab: evaluation} under the Avg. Spd. columns, where we show that the average speed of the ego-vehicle is lower in conservative environments than in aggressive environments. This variation of speed is desirable since it is reflects that ego-vehicle slows down to overtake conservative vehicles and is able to maintain a high speed in fast-moving aggressive traffic. Such adaptability results in better fuel efficiency, increased safety, and reduced likelihood of frustrations among drivers. However in default traffic, the ego-vehicle is unable to adapt its speed (Table~{\ref{tab: evaluation}}).

\section{Conclusions, Limitations, and Future Work}
\label{sec: conclusion}
We have presented a new approach for driver behavior modeling that uses the idea of vertex centrality from computational graph theory to explicitly model the behavior of human drivers in realtime traffic using only the trajectories of the vehicles in the global coordinate frame. Our approach is robust, general, and can be integrated with existing navigation methods to perform behaviorally-guided navigation. 

There are several interesting directions of future work. Our work is currently limited to straight roads. It would be useful to apply our approach to additional scenarios, including roundabouts, intersections, and merging. Another aspect of future work includes extending our approach for decision-making. While our current approach stops at modeling the driver behavior, a natural extension of our work includes combining our algorithm with motion and decision planning techniques for end-to-end self-driving.

\begin{appendices}
\section{Proving $\Vts{\tilde \beta - \beta} =  \bigO{\epsilon}$.}
\label{appendix: proof}



\begin{figure}[t]
    \centering
    \includegraphics[width = \columnwidth]{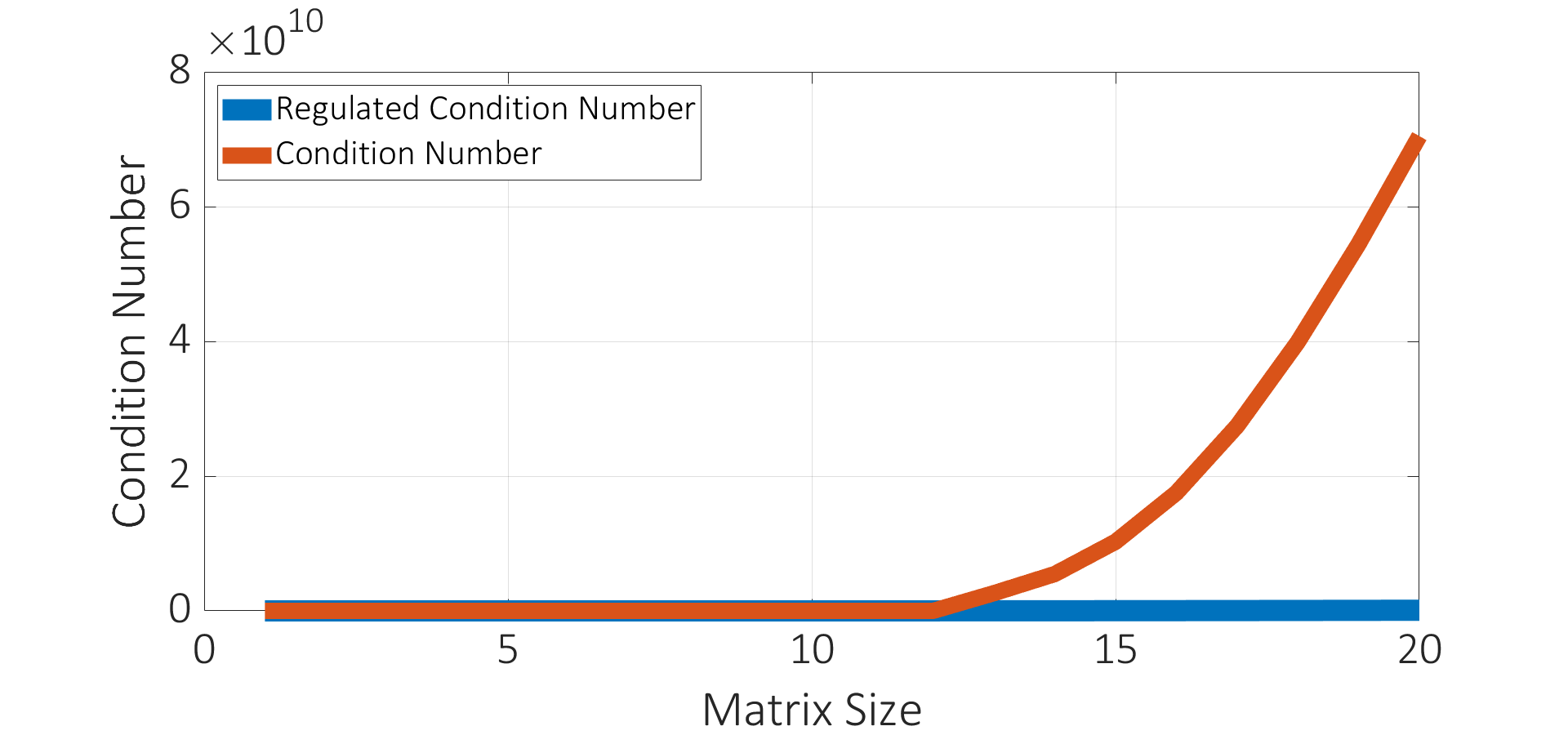}
    \caption{\textbf{Robustness to Noise:} We show that by regularizing the noisy OLS system given by Equation~\ref{eq: noisy_OLS}, we can reduce the original condition number (red curve) while at the same time upper bounding the reduced condition number (blue curve) by $\delta \longrightarrow 1$. The reduced condition number helps stabilize the noisy estimator $\tilde \beta$. }
    \label{fig: condition_number}
\end{figure}

\begin{proof}
$M$ is $T\times(d+1)$ Vandermonde matrix where $d \ll T$ is the degree of the resulting centrality polynomial. Vandermonde matrices are known to be ill-conditioned with high condition number $\kappa = \frac{\sigma_\textrm{max} \Bstrutfrac}{\sigma_\textrm{min}}$ that increases exponentially with time $T$. From the noisy system given by Equation~\ref{eq: noisy_OLS}, we have,

\begin{equation}
    \begin{split}
        \tilde\beta &= \inv{(M^\top M)} M^\top\tilde\zeta^i \\
        \tilde\beta &= \inv{(M^\top M)} M^\top(\zeta^i + \epsilon) \\
        \tilde\beta &= \beta + \inv{(M^\top M)}M^\top\epsilon \\
    \end{split}
    \label{eq: beta_betatilde_relationship}
\end{equation}

\noindent From Equation~\ref{eq: beta_betatilde_relationship}, $\Vts{\tilde \beta - \beta} = \Vts{\inv{(M^\top M)}M\epsilon}$ which can be shown to be approximately in the order $ \bigO{\kappa\epsilon}$. Therefore, the error between the true solution $\beta$ and the estimated solution in the presence of noise $\tilde \beta$, depends on the condition number $\kappa$ of the matrix $M$. A higher value of $\kappa$ implies that the trailing singular values of $M^\top M$, denoted by $\Sigma = \{ \sigma_1^2, \sigma_2^2, \ldots, \sigma_d^2 \}$ have very small magnitudes. When inverting the matrix $M^\top M$, as in Equation~\ref{eq: beta_betatilde_relationship}, the singular values are inverted (by taking the reciprocal) and are represented by $\inv{\Sigma} = \{ \inv{(\sigma_1^2)}, \inv{(\sigma_2^2)}, \ldots, \inv{(\sigma_d^2)} \}$. After the inversion, the trailing singular values now have large magnitudes, $\Vts{\inv{(\sigma_i^2)}} \gg 1$. When multiplied by $\epsilon$, these large inverted singular values amplify the error, resulting in a large value of $\Vts{\tilde \beta - \beta}$. In Figure~\ref{fig: condition_number}, it can be seen by the red curve that under general conditions, $\kappa$ increases exponentially for even small matrices (we fix $d=2$ and increase $T$ from $0$ to $20$). 

However, there are many techniques that bound the condition number of a matrix by regularizing its singular values. In our application, we use the well-known Tikhonov regularization~\cite{calvetti2003tikhonov}. Under this regularization, after the inversion operation is applied on $M^\top M$, the magnitude of the resulting inverted singular values are constrained by adding a parameter $\alpha$. The modified inverted singular values can be expressed as $\frac{\sigma_i \Bstrutfrac}{\Tstrutfrac \sigma_i^2 + \alpha^2}$. The addition of the $\alpha^2$ in the denominator keeps the overall magnitude of the inverted singular value from ``blowing up''. 
In our approach, as $M$ is fixed for every $T$, we need only search for the optimal $\alpha$ once for every $T$. Using the Tikhonov regularization, we upper bound the condition number $\kappa$ by $\delta \rightarrow 1$ (Figure~\ref{fig: condition_number}). Therefore $\Vts{\tilde \beta - \beta} = \bigO{\kappa\epsilon} = \bigO{\epsilon}$ as $\kappa \rightarrow 1$.




\end{proof}
\section{Simulation Parameters}
\label{sec: S2}
We use the Highway-Env simulator~\cite{leurent2019social} developed using PyGame~\cite{pygame}. The simulator consists of a 2D environment where vehicles are made to drive along a multi-lane highway using the Bicycle Kinematic Model~\cite{polack2017kinematic} as the underlying motion model. The linear acceleration model is based on the Intelligent Driver Model (IDM)~\cite{treiber2000congested}, while the lane changing behavior is based on the MOBIL~\cite{kesting2007general} model.

The linear acceleration model is based on the Intelligent Driver Model (IDM)~\cite{treiber2000congested} and is computed via the following kinematic equation,
\begin{equation}
    \dot v_{\alpha} = a\begin{bmatrix}1 - (\frac{v_{\alpha}}{v_0^{\alpha}})^4 - (\frac{s^*(v_{\alpha}, \Delta v_{\alpha})}{s_{\alpha}})^2\end{bmatrix}
    \label{eq: IDM_acc}
\end{equation}

\noindent Here, the linear acceleration, $\dot v_{\alpha}$, is a function of the velocity $v_{\alpha}$, the net distance gap $s_{\alpha}$ and the velocity difference $\Delta v_{\alpha}$ between the ego-vehicle and the vehicle in front. Equation~\ref{eq: IDM_acc} is a combination of the acceleration on a free road $\dot v_{free} = a[1 - (v/v_0)^{4}]$ (\textit{i.e.} no obstacles) and the braking deceleration, $-a(s^*(v_\alpha,\Delta v_\alpha)/s_\alpha)^2$ (\textit{i.e.} when the ego-vehicle comes in close proximity to the vehicle in front). The deceleration term depends on the ratio of the desired minimum gap ($s^*(v_\alpha,\Delta v_\alpha)$) and the actual gap ($s_{\alpha}$), where $s^* (v_\alpha,\Delta v_\alpha)= s_0 + vT + \frac{v\Delta v}{2\sqrt{ab}}$. $s_0$ is the minimum distance in congested traffic, $vT$ is the distance while following the leading vehicle at a constant safety time gap $T$, and $a,b$ correspond to the comfortable maximum acceleration and comfortable maximum deceleration, respectively. 

\begin{table}[t]
\caption{We show the simulation parameters that define the conservative and aggressive vehicle classes.}
\centering
\resizebox{\columnwidth}{!}{%
\begin{tabular}{cccc} 
\toprule
Model & Parameter \Tstrut & Conservative \Bstrut &   Aggressive \\
\hline
\multirow{4}{*}{IDM}& Time gap ( $T$) \Tstrut & 1.5s      & 1.2s \\
 &Min distance ($s_0$) & 5.0 $m$ & 2.5 $m$ \\
&Max comfort acc. ($a$)     & 3.0 $m/s^2$ & 6.0 $m/s^2$\\
&Max comfort dec. ($b$) & 6.0 $m/s^2$               &  9.0 $m/s^2$ \\
\midrule
\multirow{3}{*}{MOBIL}& Politeness ($p$) & 0.5     & 0\\
& Min acc gain ($\Delta a_{th}$) & 0.2 $m/s^2$ & 0 $m/s^2$ \\
& Safe acc limit ($b_{safe}$) & 3.0 $m/s^2$ & 9.0 $m/s^2$\\
\bottomrule
\end{tabular}
}
\label{tab: parameters}
\vspace{-10pt}
\end{table}

The lane changing behavior is based on the MOBIL~\cite{kesting2007general} model. According to this model, there are two key parameters when considering a lane-change:
\begin{enumerate}
    \item \textit{Safety Criterion}: This condition checks if, after a lane-change to a target lane, the ego-vehicle has enough room to accelerate. Formally, we check if the deceleration of the successor $a_\textrm{target}$ in the target lane exceeds a pre-defined safe limit $b_{safe}$:
    \begin{equation*}
        a_\textrm{target} \geq -b_{safe}
    \end{equation*}
    
    \item \textit{Incentive Criterion}: This criterion determines the total advantage to the ego-vehicle after the lane-change, measured in terms of total acceleration gain or loss. It is computed with the formula,
    
    \begin{equation*}
    \tilde{a}_\textrm{ego} - a_\textrm{ego} + p(\tilde{a}_n - a_n + \tilde{a}_o - a_o) > \Delta a_{th}
    \end{equation*}
    where $\tilde{a}_\textrm{ego} - a_\textrm{ego}$ represents the acceleration gain that the ego-vehicle would receive after to the lane change. The second term denotes the total acceleration gain/loss of the immediate neighbours (the new follower in the target, $a_n$, and the original follower in the current lane, $a_o$) weighted with the politeness factor, $p$. By adjusting $p$ the intent of the drivers can be changed from purely egoistic ($p=0$) to more altruistic ($p=1$). We refer the reader to~\cite{kesting2007general} for further details.
\end{enumerate}

\noindent The lane change is executed if both the safety criterion is satisfied, \textit{and} the total acceleration gain is more than the defined minimum acceleration gain, $\Delta a_{th}$.

Additionally, the desired velocity $v_0$ is set to $25$ meters per second and $40$ meters per second for the conservative and aggressive vehicle classes, respectively. Finally, the desired velocities for the conservative vehicles were uniformly distributed with a variation of {$\pm$10\%} to increase the heterogeneity in the simulation environment.

\end{appendices}

\section*{Acknowledgement}
This work was supported in part by ARO Grants W911NF1910069 and W911NF1910315, Semiconductor Research Corporation (SRC), and Intel.



\bibliographystyle{ieee}
\bibliography{refs}
%
\begin{IEEEbiography}[{\includegraphics[width=1in,height=1.25in,clip,keepaspectratio]{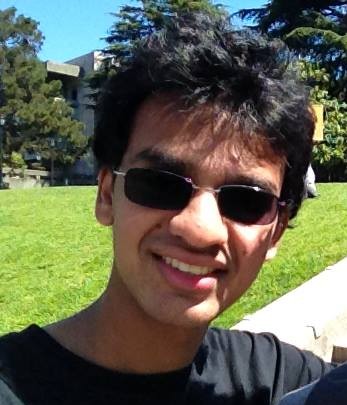}}]
 {Rohan Chandra}
is a PhD student in computer science at University of Maryland, College Park, USA (UMD). He received a Bachelors degree in ECE from Delhi Technological University, New Delhi, India, and a Masters degree in computer science from UMD. His research interests include trajectory prediction, behavior modeling, and navigation in autonomous driving.
\end{IEEEbiography}

\begin{IEEEbiography}[{\includegraphics[width=1in,height=1.25in,clip,keepaspectratio]{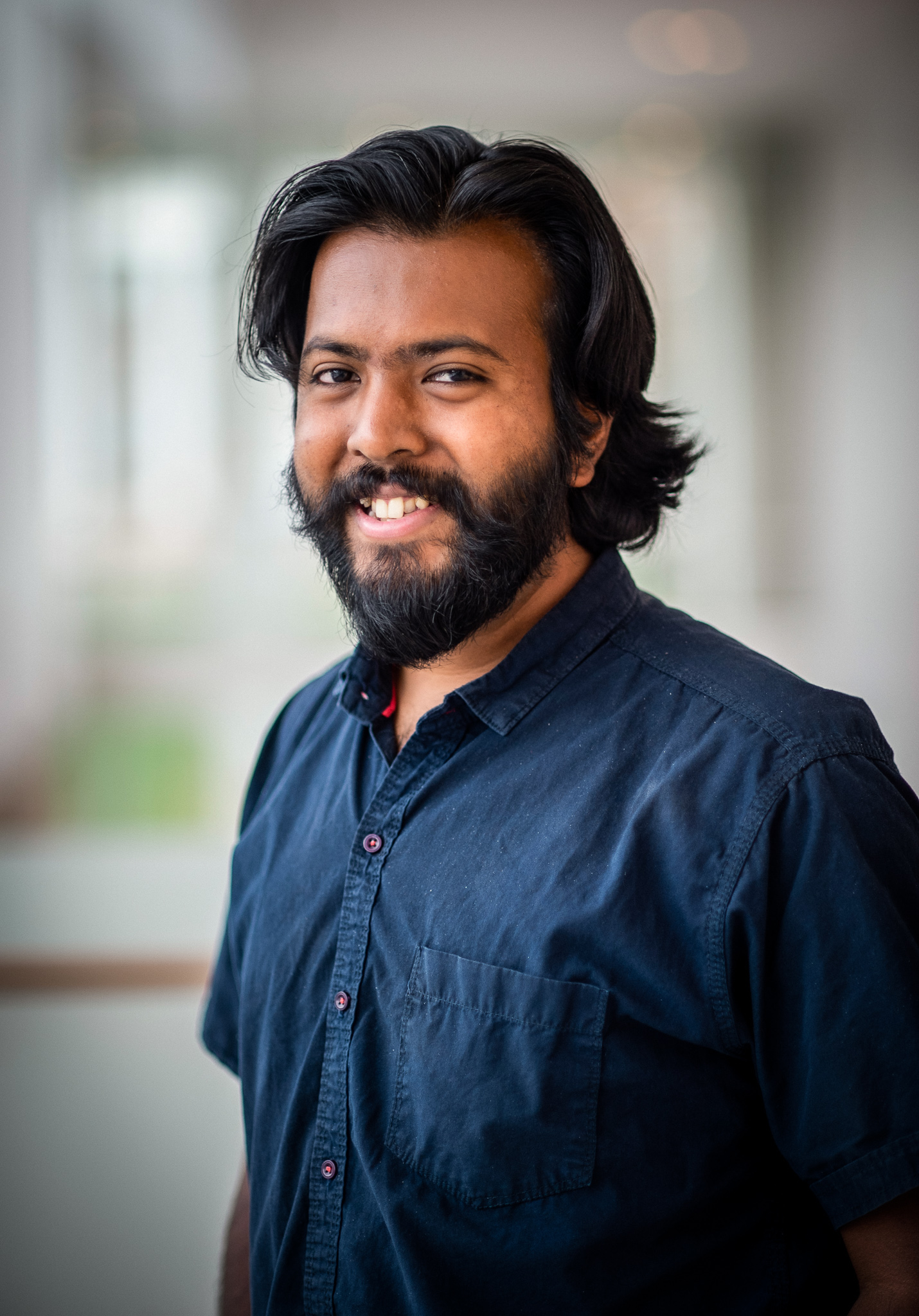}}]{Aniket Bera}

is an Assistant Research Professor in the Department of Computer Science with affiliated appointments with the Brain and Behavior Institute and Maryland Robotics Center. Prior to this, he was a Research Assistant Professor at the University of North Carolina at Chapel Hill, where he also received his Ph.D. in 2017. His core research interests are in Human Modeling and Simulation, Affective Computing, and Virtual Reality. He is a faculty with the GAMMA group and has advised and co-advised over 20 MS and Ph.D. students. He has authored over 50 papers and his work has won multiple awards at top VR conferences.  His research involves novel combinations of methods and collaborations in affective computing, computer graphics, computational psychology, machine learning, and social robotics to develop real-time computational models to learn and simulate human-like agents with expressive behaviors. Dr. Bera has previously worked in many research labs, including Disney Research and Intel Labs, and his work has been featured in many media outlets including Forbes, WIRED, FastCompany, etc. A more detailed description can be found at http://cs.umd.edu/~ab.
\end{IEEEbiography}

\begin{IEEEbiography}[{\includegraphics[width=1in,height=1.25in,clip,keepaspectratio]{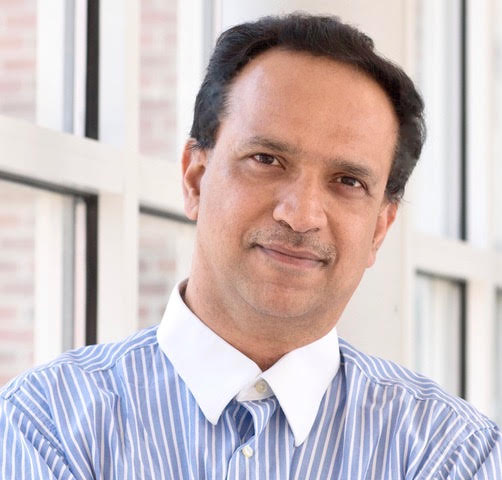}}]{Dinesh Manocha}

is the Paul Chrisman-Iribe Chair in Computer Science \& Electrical and Computer Engineering and Distinguished University Professor at University of Maryland College Park. His research interests include virtual environments, physically-based modeling, and robotics.He has published more than 600 papers \& supervised 40 PhD dissertations. He is an inventor of 10 patents, which are licensed to industry.  He is a Fellow of AAAI, AAAS, ACM, and IEEE, member of ACM SIGGRAPH Academy, and Bézier Award recipient from Solid Modeling Association. He received the Distinguished Alumni Award from IIT Delhi the Distinguished Career in Computer Science Award from Washington Academy of Sciences.  He was a co-founder of Impulsonic, a developer of physics-based audio simulation technologies, which was acquired by Valve Inc in November 2016. 
\end{IEEEbiography}
\vfill
\vfill
\vfill
\vfill
\vfill
\vfill
\vfill
\vfill
\vfill
\vfill
\vfill
\vfill
\vfill
\vfill
\vfill
\vfill
\vfill
\vfill
\vfill
\vfill
\vfill
\vfill






\end{document}